# “Reasoning” with Rhetoric: On the Style-Evidence Tradeoff in LLM-Generated Counter-Arguments

**Preetika Verma, Kokil Jaidka, Svetlana Churina**

Centre for Trusted Internet & Community, National University of Singapore
preetikaverma08@gmail.com, jaidka@nus.edu.sg

## Abstract

Large language models (LLMs) play a key role in generating evidence-based and stylistic counter-arguments, yet their effectiveness in real-world applications has been underexplored. Previous research often neglects the balance between evidentiality and style, which are crucial for persuasive arguments. To address this, we evaluated the effectiveness of stylized evidence-based counter-argument generation in Counterfire, a new dataset of 38,000 counter-arguments generated by revising counter-arguments to Reddit’s ChangeMyView community to follow different discursive styles. We evaluated generic and stylized counter-arguments from basic and fine-tuned models such as GPT-3.5, PaLM-2, and Koala-13B, as well as newer models (GPT-4o, Claude Haiku, LLaMA-3.1) focusing on rhetorical quality and persuasiveness. Our findings reveals that humans prefer stylized counter-arguments over the original outputs, with GPT-3.5 Turbo performing well, though still not reaching human standards of rhetorical quality nor persuasiveness indicating a persisting style-evidence tradeoff in counter-argument generation by LLMs. We conclude with an examination of ethical considerations in LLM persuasion research, addressing potential risks of deceptive practices and the need for transparent deployment methodologies to safeguard against misuse in public discourse.

## Introduction

Political argumentation in online platforms has increasingly become a focal point for research in computational social science, particularly in the context of debates and political polarization. While counter-arguments are essential for fostering debate and promoting critical thinking, several key gaps persist in the literature. First, much of the existing research focuses on generating counter-arguments using rule-based or data-driven methods (Hidey and McKeown 2019), but these studies are from the pre-Large Language Model (LLM) era. Second, they overlook the role of stylistic elements, which are crucial for persuasion and attitudinal change in real-world discussions. Third, although the presentation of evidence is vital for credibility (Pornpitakpan 2004), previous work has not sufficiently examined how style—such as justification and reciprocity—affects the reception and impact of these arguments. As a result, current models may generate factually accurate counter-arguments that lack the rhetorical finesse necessary to influence opinions, especially in politically charged environments. In sum, a substantive gap exists in political counter-argument generation that infuses style with evidence.

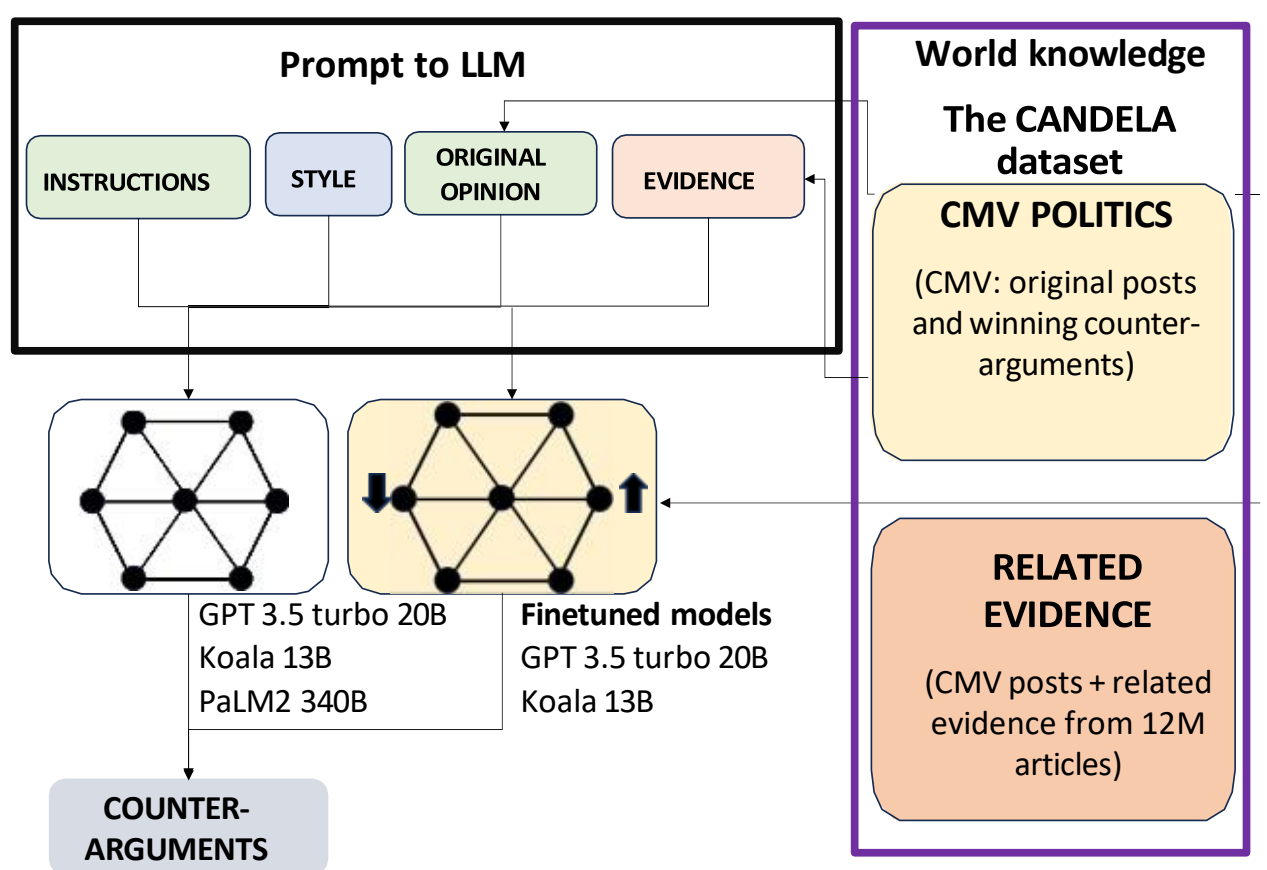

Figure 1: Experimental framework for generation.

A solution may be by following the approaches that realize different deliberative styles for successful persuasion (Jaidka 2022; Jaidka, Zhou, and Lelkes 2019; Papakyriakopoulos, Engelmann, and Winecoff 2023). In this study, we evaluated the effectiveness of stylized evidence counter-arguments vs. non-stylized ones, on the base- and fine-tuned large language models. We used prompts from posts in the ChangeMyView sub-Reddit and ultimately curated the Counterfire dataset to better understand how evidence and style are synthesized in political arguments and their relative preference by human evaluators. The Counterfire dataset, which integrates both evidence and style, addresses the lack of datasets designed for stylistically rich counter-argument generation. The dataset’s use of argument triplets—original posts, evidence, and counter-arguments—provides a unique foundation for understanding how style interacts with content to shape human preferences. The code and dataset are available at https://github.com/Preetika764/Style_control/.

We make three key contributions:

- A retrieval-augmented generative pipeline that produces stylized counter-arguments grounded in political communication theory
- An argument triplets dataset for studying fine-grained style control for political discourse
- Human preference and persuasive labels to model the tradeoff between evidence integration and argument quality for researchers in human-computer interaction.

Our approach to refine LLM-generated counter-arguments applies insights from political communication, particularly the distinctions between analytical and social arguments. While analytical arguments rely heavily on justification and evidence, social arguments emphasize reciprocity and respect in dialogue (Rinke 2015). We have proposed a fine-grained evaluation framework validates the output, analyzes crowd-sourced annotations on the quality and relative persuasiveness of one style over the other, and offers discursive insights by comparing rhetorical and discursive markers. The findings suggest that while humans prefer a stylized counter-argument over a neutral one, the limited discursive range of generated outputs may explain why current state-of-the-art would still lose out to human counter-arguments in the short term.

The broader impact of our work beyond political discourse on social media concerns domains where persuasive communication is critical. For instance, our findings on evidence integration and stylistic elements in legal reasoning could enhance argument mining from case law and help generate more persuasive legal briefs that balance analytical rigor with rhetorical effectiveness (Ashley and Walker 2013). For collaborative decision-making in organizations, our framework for distinguishing between analytical and social arguments could improve how AI systems support group deliberation by adapting argument styles to different stakeholders and contexts(Bose, Reina, and Marshall 2017). The demonstrated impact of style on argument reception also has implications for content moderation. Rather than simply flagging problematic content, moderation systems could leverage our approach to generate constructive counter-narratives that acknowledge user perspectives while encouraging more nuanced discourse(Sinem Tekiroglu, Chung, and Guerini 2020). In educational contexts, our findings about the relationship between evidence presentation and rhetorical style could inform writing assistance tools that help students develop more effective argumentative essays by suggesting improvements to both logical structure and persuasive elements(Woods et al. 2017).

## Background

Our work applies a political communication lens to the problem of evidence-based counterargument generation. First, political communication theory on the discursive style of deliberative discussions particularly the work of Rinke (2015), views social media platforms as spaces for 'internal reasoned dissent,' where users engage with diverse viewpoints in mediated deliberation. This framework categorizes arguments into two broad types: analytical arguments, such as Justification, which is focused on rationality through the presentation of evidence, and social arguments such as Reciprocity, which prioritizes interpersonal engagement in dialogue (Esteve Del Valle, Sijtsma, and Stegeman 2018). These distinctions guide our investigation into how well LLMs balance these argumentative styles when generating counter-arguments.

The stylistic definitions and measurements adopted for our research are grounded in the CLAPTON corpus (Jaidka 2022), a dataset that provides insights into argumentation styles used in Reddit's ChangeMyView platform. CLAPTON's labels suggest that Reddit users tend to favor the Justification (30%) and Reciprocity (25.8%) styles in their counter-arguments, while Constructiveness (6.6%) is less frequently employed. The CLAPTON dataset, therefore, motivated us to evaluate whether LLM-generated political counter-arguments would fare better if they applied similar strategies while also offering a benchmark for us to validate the outputs on ground truth data.

### Argument Generation Frameworks

LLMs have demonstrated strong performance across various downstream tasks, but prior work on applying them to generate counter-arguments is scarce. Generating factual and convincing counter-arguments still remains particularly challenging due to the need for external information, such as through incorporating extra steps for information retrieval and coherence. Retrieval-augmented generation (RAG) methods, such as GRAGIN (Su et al. 2024), dynamically adjust retrieval based on the evolving information needs of the LLMs, enhancing evidence integration for complex arguments. For instance, methods like M-RAG (Wang et al. 2024) improve precision by partitioning memory into smaller, focused units to enhance retrieval accuracy while potentially neglecting argumentative intent as a potential objective. However, evidence alone does not guarantee persuasive power. Our work prioritizes modifying and evaluating the impact of stylistic elements on how evidence is presented. Therefore, we have a static retrieval component that uses the outputs from the retrieval-augmentation module developed for Candela, a framework developed by Hua, Hu, and Wang (2019) that uses a seq2seq architecture to construct coherent arguments out of retrieved passages.

In summary, prior approaches treat argument generation as primarily an information retrieval challenge, overlooking the crucial role of how that evidence is presented and prioritizing factual accuracy over persuasive presentation (e.g., CB-RAG). We address the gap of combining evidence with stylistic adjustments, contributing to a better understanding of how style shapes the effectiveness of generated arguments. In comparison to systems like CBR-RAG (Wiratunga et al. 2024) that extend retrieval to case-based reasoning for legal contexts, ensuring coherence and factual accuracy, we emphasize the importance of stylistic elements such as tone, engagement, and respect in persuasive argumentation. This ensures both factual grounding and stylistic control.

### Controllable Text Generation

Recent approaches to controllable text generation has often focused on sentiment control, with various techniques surveyed in Lorandi and Belz (2023). Plug-and-play language models allow for control without retraining (Dathathri et al. 2019), yet the goal is often simply to infuse a domain vocabulary. Similarly, approaches for controlled generation via language model arithmetic (Dekoninck et al. 2023) and recent improvements in position-aware decoding for controlling output style (Gu et al. 2022) focus mainly on fine-grained control of text content. In contrast, our work focuses on macro-level style, synthesizing the content while grounded in authors' intent.

Prior approaches do not capture the higher-level stylistic elements crucial for persuasive argumentation, such as the deliberative dimensions essential for effective political discourse. Our work, that focuses on the authors' rhetorical intent, is closer to the work on generating stylized text with LLMs in areas such as readability (Collins-Thompson 2014), formality (Chhaya et al. 2018), and politeness (Althoff, Danescu-Niculescu-Mizil, and Jurafsky 2014); yet, none of these approaches consider the deliberative dimensions that are of interest to us. Our goals are also similar to prior work in domain-specific controllable text generation for scientific summarization (Ding et al. 2023) yet we apply it to political discussions. For instance, prior research on counter-argument generation (El Baff et al. 2020; Al Khatib, Morari, and Stein 2020) is divorced from the study of audience preferences effects as well the macro-level style features in the LLM era, such as styles for deliberation in political discussions. Our study explicitly models and controls deliberative dimensions of argumentation. This includes parameters for the degree of reciprocity shown, and the integration of evidence through justification.

Ultimately, our approach to persuasive political argument generation represents a fundamental shift from previous work on counter-argument generation, which has largely ignored the intersection of evidence, style, and audience preferences in the era of large language models. Beyond the evaluation frameworks proposed in prior work for long-text generation (Goyal, Li, and Durrett 2022), our unique objective required us to develop a new evaluation framework to assess stylized argumentative discourse, validating style, evidence, quality, discourse, and persuasiveness of the generated text.

## Method

Figure 1 illustrates our framework, which uses facts from seq2seq intermediate outputs to create domain-injected prompts, generating relevant, logical, and grammatical counter-arguments from off-the-shelf and fine-tuned LLMs. Our method involved the following steps:

- **Retrieval of Counter-Evidence**: We used the retrieval system from Hua, Hu, and Wang (2019) to obtain credible counter-evidence for Original Posts from the Change My View (CMV) dataset.
- **Zero-Shot Prompting**: We applied zero-shot style-focused prompts to generate evidence-based counter-arguments from CMV posts using off-the-shelf LLMs prompted with the original post, evidence, and possibly a style directive. The prompts and styles are listed in Table 1.

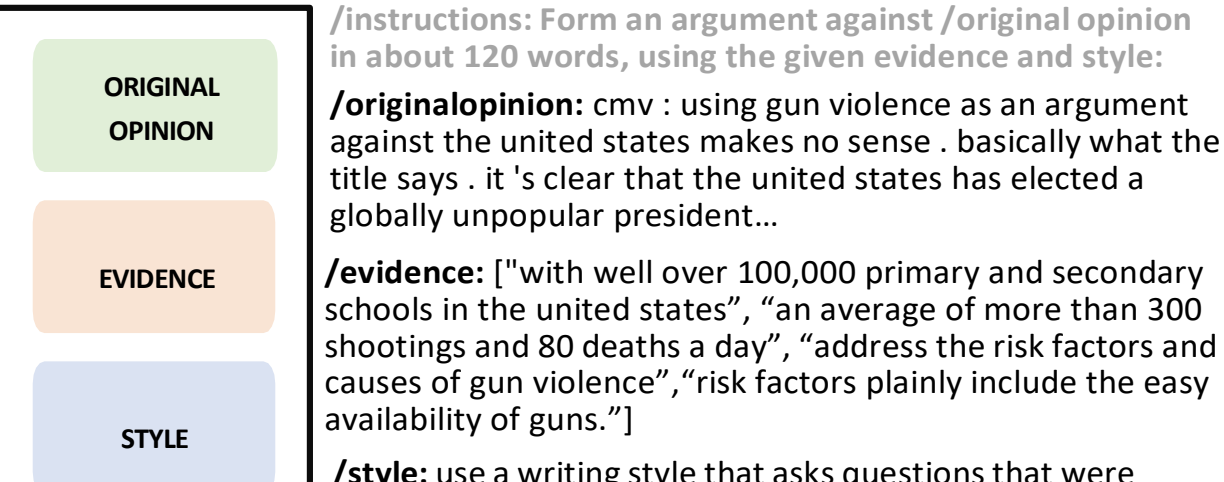

Figure 2: Example prompt for generating a reciprocal counter-argument.

| Style | Prompt |
|---|---|
| Plain | Use a writing style that focuses on using the evidence and being convincing. |
| Reciprocity | Use a writing style that asks questions designed to elicit opinions or information from the user. |
| Justification | Use a writing style that focuses on fact-reporting or fact-checking, finding common ground, and providing personal or statistical evidence with references. |

Table 1: The three variants of the style specifications added to the LLM prompt, adapted from the original definitions (Jaidka 2022).

- **Fine-Tuning**: We fine-tuned GPT-3.5 Turbo and Koala-13B with the original post and Candela-retrieved evidence on 70,000 observations from the CMV dataset (distinct from the prompts used for counter-argument generation), to improve the integration of evidence, and then performed zero-shot prompting.

## Data Collection

Counterfire [1] comprises inputs and their human-written outputs sampled from the Candela dataset, paired with LLM-generated outputs in three distinct styles. The data collection process was conducted in two phases: zero-shot prompting and fine-tuning techniques.

- **Baseline**: The primary baseline was established using the Candela dataset (Hua, Hu, and Wang 2019), which includes 70,000 English posts and counter-arguments from r/ChangeMyView, paired with deduplicated and ranked evidence from a database of 12 million articles, including sources like Wikipedia, Washington Post, New York Times, Reuters, and The Wall Street Journal. An additional random sample of 2000 rows of posts and evidence was selected for further analysis.
- **Generating Stylized Counter-Arguments**: Six off-the-shelf and fine-tuned LLMs were prompted with the original posts and evidence from the Candela dataset, incorporating stylistic variations based on prior work (Jaidka

[1] A preprint under the title "Auditing Counterfire: Evaluating Advanced Counter-argument Generation with Evidence and Style" was previously published on arxiv and has since been updated

**Original opinion:** parents who refuse vaccination of their children must sign a form of accountability so if their child dies from medical complications (...)
**Evidence:** ["about 1 percent of u.s. parents get exemptions for their children , but the figure is higher in some areas","the fear is that parents who don't vaccinate or who delay vaccinating their children put their own kids as well as others at risk for developing preventable disease , such as whooping cough"]

**Candela output:** If you want to change your view, do you really think we should force them to do things they don't want to do? (...)

**GPT-3.5 turbo output with no style:** Firstly, it is not always easy to determine the exact cause of a child's illness or death, and therefore it may not be fair to blame it solely on the lack of vaccination. Secondly, some parents may not have access to vaccinations in their area or cannot afford them, and punishing them would be unfair. (...)

**GPT-3.5 turbo output with reciprocity:** While I understand the frustration towards parents who refuse to vaccinate their children, do you think holding them accountable for their child's potential death is the best approach? What if the child had an allergic reaction to the vaccine or experienced complications that were rare but severe? How would this legal contract work in those situations? (...)

**GPT-3.5 turbo output with justification:** The proposed idea to hold parents accountable for not vaccinating their children is misguided and unfair. First and foremost, compulsory vaccination violates personal freedom. Parents have the right to make decisions for their children, including when it comes to medical procedures. (...)

Figure 3: An example input statement and the generated counter-arguments for the CMV dataset from GPT-3.5 turbo.

2022). A set of prompts without style was also created for comparison. Figure 2 shows a sample prompt for generating a reciprocal counter-argument, which shows how the original opinion and the related evidence were provided to the LLM with or without a request for a stylized counter-argument. The style instructions are listed in Table 1. We generated 32,000 counter-arguments using 2000 inputs across 3 styles with 5 LLMs.

- **LLM Details**: The six LLMs we used for our experiments are reported roughly in chronological order of release:
  - **GPT-3.5 Turbo**: This model was used for generating counter-arguments with style control, leveraging the same input passages as the baseline for consistency. Example counter-arguments generated by GPT-3.5 Turbo are shown in Figure 3.
  - **Koala-13B**: Koala-13B, fine-tuned from LLaMA, was trained on high-quality, deduplicated datasets to improve its performance on counter-argument generation.
  - **Fine-Tuned Variants of GPT-3.5 Turbo and Koala-13B**: GPT-3.5 Turbo and Koala-13B were fine-tuned on the dataset of 70,000 input-counter-argument pairs using specialized settings for each model. We ensured that the input dataset for zero-shot prompting was distinct from the dataset used for fine-tuning. Specifically, the dataset for zero-shot prompting consisted of 2000 posts from the CMV dataset, while the fine-tuning dataset included a separate set of 70,000 posts. This distinction prevented any overlap between the data used for model training and evaluation and ensured the integrity of the fact integration validation results.
  - **GPT-4o**: This model have been used to generate conuter-arguments with style control. This model introduces improved reasoning capabilities and enhanced factual consistency, making it particularly suited for tasks requiring nuanced counter-argument generation.
  - **LLaMA-3.1-8B**: LLaMA-3.1 performs well in fine-tuned domains and diverse input handling.
  - **PaLM-2 Text-Bison**: PaLM-2, known for its multilingual and reasoning capabilities, was initially tested but excluded from fine-tuning due to its tendency to generate supporting arguments instead of counter-arguments.
  - **Claude Haiku**: Designed for concise, ethical text generation, Claude Haiku excels at producing clear, balanced arguments in fewer tokens.

  Ultimately, we did not fine-tune all six of the LLMs, as (except in the case of PaLM-2, where generation was inaccurate) we observed no perceptible improvement in counter-argument generation, as evidenced in the marginal differences between the non- and fine-tuned models in human evaluations, reported in Table 6 of the Appendix. Furthermore, the findings related to our primary focus on the comparative effectiveness of stylized counter-arguments follow the trends from the base models.

The fine-tuning process, including hyperparameters and training loss, is provided in the Appendix. Example counter-arguments generated by GPT-3.5 Turbo are reported in Figure 3. For further details on the generation process and evaluations, please refer to the *Results* section. The hyperparameters and fine-tuning configurations for the LLMs are provided in the Appendix. PaLM-2 was excluded from fine-tuning due to its lack of adherence to prompts, such as its tendency to generate supporting arguments instead of counter-arguments.

## Analyses

Our analytical approach comprised human quality assessments of all the outputs from all the models. Subsequently, our analyses drilled down on GPT-3.5 turbo outputs because they were the most preferred by humans in our quality evaluations (even more than the newer GPT-4o models). On this subset, we performed a validation task, rhetorical analysis, and a human preference anlaysis. Our procedure is described below.

- **Human Quality Assessments**: Collected human assessments across four quality dimensions: content, grammar, logic, and overall effectiveness. The detailed instructions for the crowdsourced quality assessments are provided in the Appendix.

- **Validation Tasks**: Performed validation tasks to audit the LLMs' ability to adhere to instructed prompts on counter-arguments generated by GPT-3.5 Turbo:
  - **Fact Integration**: Assessed whether the LLMs effectively incorporated the provided evidence into the generated counter-arguments, using metrics like BERTScore (Zhang et al. 2019) and ROUGE-1.
  - **Style Validation**: Evaluated whether the outputs reflected the expected discussion style (justification or reciprocity) through both automatic and human evaluations.
- **Rhetorical Analysis**: Conducted an automatic rhetorical analysis to compare the GPT-3.5 turbo counter-arguments against the baseline and human arguments on readability, rhetorical intent, and discourse. These are reported in the appendix.
  - **Readability Metrics**: Calculated Flesch-Kincaid, Flesch Reading Ease, Gunning Fog, and Smog Index using the textstat package.
  - **Rhetorical Intent**: Characterized the counter-arguments by analyzing rhetorical moves, including alignment, authority, and persuasion, using the AAWD taxonomy of argumentation (Bender et al. 2011).
  - **Discourse Analysis**: Used Python's convokit package to identify various persuasive moves, applying previously established methods in online persuasion research (Danescu-Niculescu-Mizil et al. 2013).
- **Human Preference Analysis**: Besides the annotations for counter-argument effectiveness collected as part of the human quality assessments (described in the first paragraph of this subsection), we also conducted a survey to evaluate preferences for GPT-3.5 counter-arguments as compared to human-written counter-arguments. The survey collected 10,000 rankings of counter-arguments through Amazon Mechanical Turk. Workers were instructed to rank counter-arguments posted on Reddit in the order from most to least preferred.

# Results

## Human Quality Assessments

We conducted a comprehensive manual evaluation of the quality of the generated counter-arguments to identify the most-preferred LLM. We launched an Amazon Mechanical Turk task to obtain eight annotations per argument on facets such as content, grammar, logic, and overall effectiveness (Goyal, Li, and Durrett 2022), on a sample of 100 counter-arguments generated across the different models. In total, 2,400 counter-arguments were generated from the eight LLMs (100 inputs x 3 styles x 8 LLMs). For each counter-argument, eight unique human annotators evaluated the outputs across five quality dimensions. This resulted in a total of 19,200 annotations collected through Amazon Mechanical Turk. Annotators were paid $0.10 per HIT for an average pay of $7.25 per hour, based on the self-reported wages reported on Turkerview.[2] To ensure quality annotations, we recruited human evaluators with a minimum of 95% task approval rating and prior experience in content evaluation tasks. Annotations were aggregated by calculating the mean score for each quality dimension per argument. Figure 4 shows the human evaluation of counter-arguments generated by various models, mainly focusing on the newer and older variants of the OpenAI GPT and LLaMA models and the Claude haiku model. Each boxplot represents the distribution of scores for content, grammar, and logic. The whiskers denote the range of scores, and the outliers are represented by dots. Different colors and box styles correspond to the specific LLM models and style prompts used.

[2]https://turkerview.com/requesters/

The scores for Justification variants are significantly higher than the Reciprocity variant for the GPT-3.5 Turbo and GPT-4o models for the logic and content dimensions. Among older models, the counter-arguments with no style also tend to perform consistently better than reciprocity across different metrics, and the difference is statistically significant among GPT-3.5 Turbo models. However, the trend appears to be changing with the newer models, where reciprocity counter-arguments are significantly better than the former. These findings suggest that while stylistic variations can enrich counter-arguments, they may contradict the expectations of quality counter-arguments depending on the version of the LLM used. The complete results are reported in the supplementary materials, where we also observed no significant advantage of fine-tuning LLMs. All models significantly outperformed counter-arguments generated by the baseline system, Candela (Hua, Hu, and Wang 2019).

## Validation

We performed two validation tasks to audit the ability of LLMs to adhere to the instructed prompts: (a) Fact integration, assessing the incorporation of evidence, and (b) Style validation, evaluating whether the outputs reflect the expected discussion style.

**Fact Integration Validation** For fact integration validation, we analyzed whether our prompts effectively guided the LLMs to incorporate the provided evidence into the generated counter-arguments. This involved comparing the similarity and absolute overlap of evidence with the outputs, using metrics such as BERTScore (Zhang et al. 2019) and ROUGE-1 (Lin 2004), because they measure both semantic similarity (BERTScore) and token overlap (ROUGE-1) between the generated counter-arguments and the evidence provided. While BERTScore captures the ability of models to paraphrase evidence effectively, ROUGE-1 is sensitive to exact token overlap, offering insights into how closely the generated text aligns with the original source.

Table 2 presents a comparative evaluation of different large language models (LLMs) and their prompting strategies using various automatic evaluation metrics. The first part of the table compares the fact integration capabilities of GPT-3.5 Turbo, Koala-13B, and Palm-2 using BERTscore (F1 value) and ROUGE-1 (Recall), suggesting that there is not much to choose between the three based on adherence to content. Yet, note that the average BERTScore F1 value across the three LLMs is 0.725, and the average ROUGE-1 recall is 0.313, suggesting that, as can be expected, while LLMs are effective at paraphrasing evidence into counter-arguments, they have a lower absolute overlap in the words used.

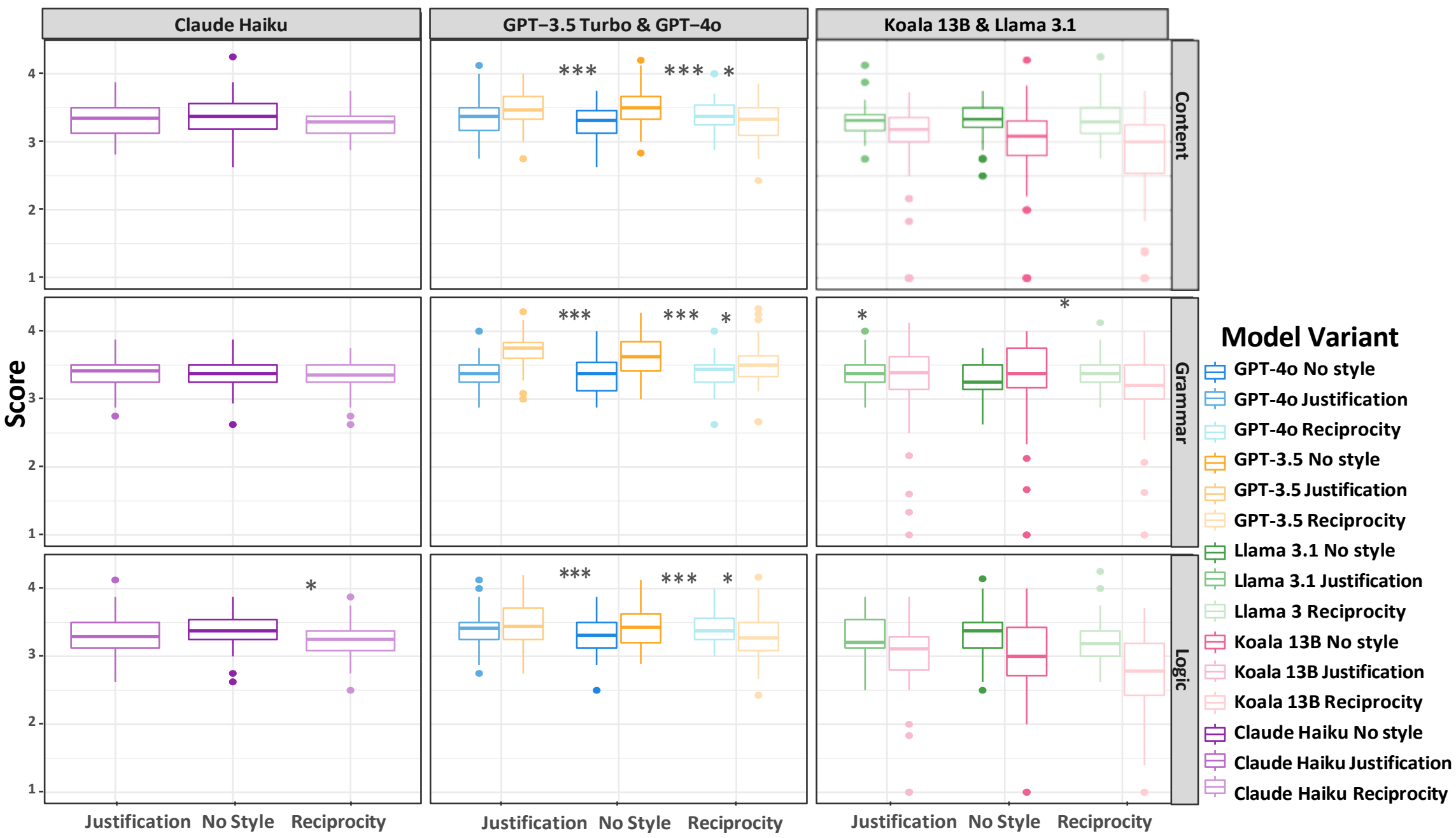

Figure 4: Results from the human evaluation of generated counter-arguments. The justification counter-arguments outperform the other variants consistently (and significantly in the case of GPT-3.5 Turbo and GPT-4o) on Content, Grammar, and Logic. Reciprocity counter-arguments from GPT-4o models significantly outperform the no-style counterarguments on all dimensions, while those from LLaMA-3.1 outperform Koala-13B on Grammar. Asterisks (*) denote that the model variant significantly outperformed others in the same group based on pairwise comparisons using a Bonferroni-corrected statistical significance (*** :$p < 0.001$; ** :$p < 0.01$; *:$p < 0.05$).

| Metric | GPT-3.5 Turbo | Koala-13B | PaLM2 |
|---|---|---|---|
| **Comparing LLMs** | | | |
| BERTscore (F1 Value) | 0.7312 | 0.7271 | 0.7175 |
| ROUGE-1 (Recall) | 0.3556 | 0.3631 | 0.3103 |

| Metric | Candela | GPT-3.5 Turbo No Style | GPT-3.5 Turbo Justification | GPT-3.5 Turbo Reciprocity |
|---|---|---|---|---|
| **Comparing Prompting Strategies** | | | | |
| ROUGE-1 | 0.24 (0.07) | **0.33 (0.07)** | 0.17 (0.06) | 0.17 (0.06) |
| ROUGE-2 | 0.03 (0.03) | **0.10 (0.06)** | 0.02 (0.02) | 0.01 (0.02) |
| ROUGE-L | 0.21 (0.06) | **0.29 (0.07)** | 0.15 (0.05) | 0.14 (0.04) |
| BLEU | 0.00 (0.01) | **0.06 (0.06)** | 0.00 (0.01) | 0.00 (0.01) |

Table 2: Automatic evaluation metrics and fact integration scores for different models.

Therefore, the second part of the table evaluates the content and style adherence of just GPT-3.5 Turbo, the best performing model according to human quality annotations, under different prompting strategies, compared to a baseline (Candela). The metrics used are ROUGE-1, ROUGE-2, ROUGE-L, and BLEU. The "No Style" prompt for GPT-3.5 Turbo achieves the highest scores across all the metrics, suggesting that no styling instructions allow better content adherence that outperforms the Candela baseline and other style-specific prompts. This indicates that while stylistic prompts can tailor the generated content to specific styles, they might compromise the overall content quality and factual accuracy.

**Style Integration Validation** For style validation, we examined whether the LLMs could integrate the expected style into the outputs. This was conducted using both automatic and human methods. Automatically, we fine-tuned OpenAI Ada models on the CLAPTON dataset (Jaidka 2022) to label the presence of justification and reciprocity in the generated outputs. Results for the style validation are reported in Table 3. The first half of the table reports an automatic evaluation of style, where PaLM-2 demonstrates the highest accuracy (0.49), significantly outperforming GPT-3.5 Turbo (0.17) and Koala-13B (0.09). While PaLM-2 excels in integrating the reciprocity style, GPT-3.5 Turbo is more effective in incorporating the justification style. For justification adherence, GPT-3.5 Turbo leads with an accuracy of 0.42, followed by Koala-13B (0.26) and PaLM-2 (0.22).

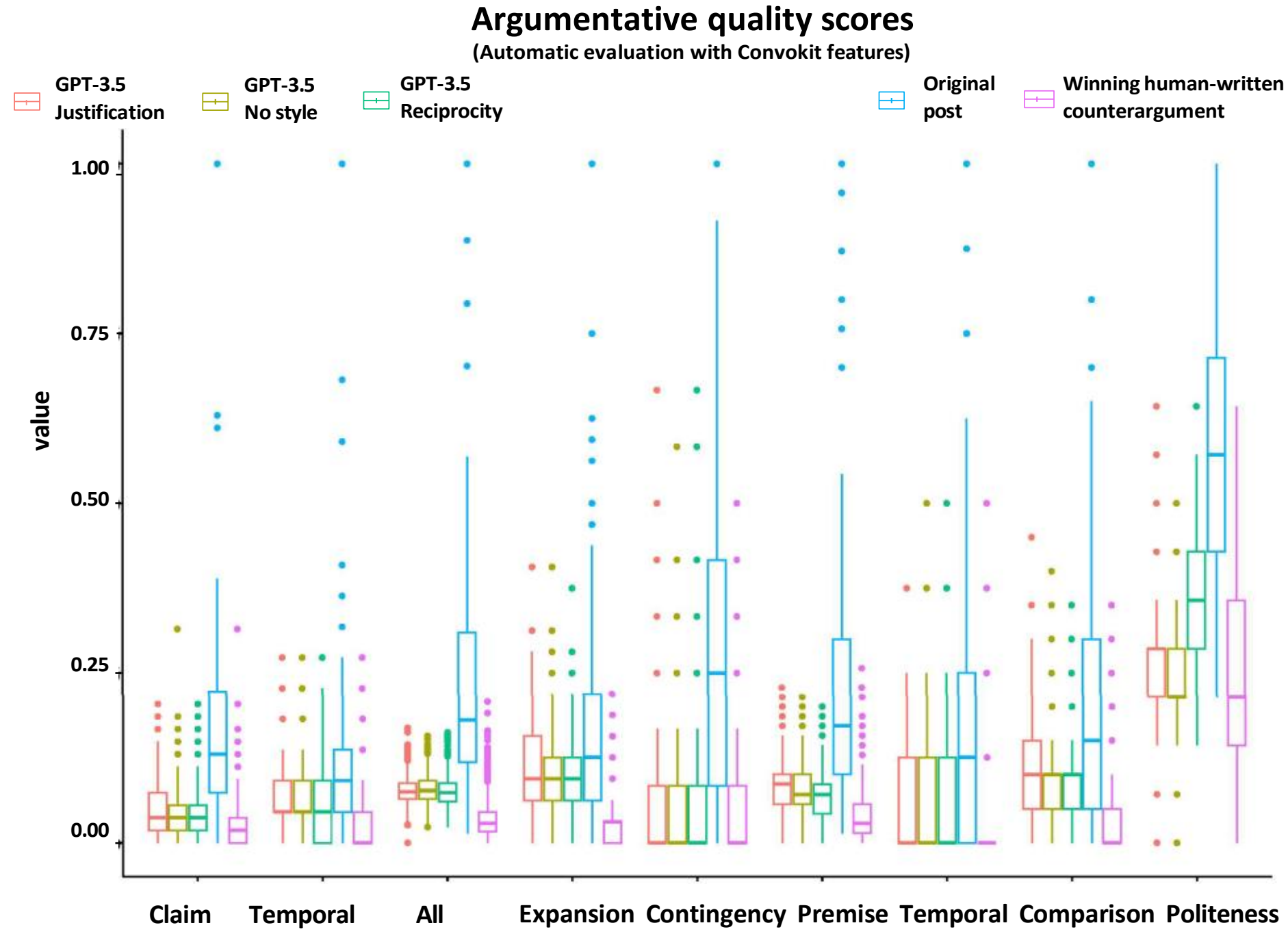

Figure 5: Results from the automatic evaluation of argumentation using Convokit highlights the discursive richness and low politeness of human counter-arguments.

| Style Integration | | | |
|---|---|---|---|
| **Style** | **GPT-3.5 Turbo** | **Koala-13B** | **PaLM2** |
| **Automatic Evaluation: Style (Accuracy)** | | | |
| Reciprocity | 0.17 | 0.09 | **0.49** |
| Justification | **0.42** | 0.26 | 0.22 |
| **Human Evaluation: Style Integration** | | | |
| **Style** | **$\theta$ (Inter-annotator Accuracy)** | | |
| Reciprocity | 0.9682 | | |
| Justification | 0.7680 | | |

Table 3: Evaluation of style integration through (a) automatic labeling with fine-tuned models, and (b) crowdsourced human labels. $\theta$ is the average annotator accuracy across true-positives and negatives (Passonneau and Carpenter 2014).

Human assessments of style (on GPT-3.5 Turbo outputs, which were evaluated as the highest quality among all variants) were obtained through an Amazon Mechanical Turk task, where a random sample of 100 counter-arguments received five annotations each on whether they exhibited cues of justification and reciprocity. We then measured their $\theta$, which allows inter-annotator agreement calculation on a five-point scale with chance-based metrics (Passonneau and Carpenter 2014) and has been widely applied in other computational linguistic tasks (Davani, Díaz, and Prabhakaran 2022; Jaidka, Ahuja, and Ng 2024). Unlike chance-based metrics, which have wide error bounds, model-based measures consider the actual categories of items in the corpus and the prevalence of each label to report the accuracy of reporting the correct answer through an expectation maximization approach. The average inter-annotator reliability was $\theta = 0.768$ for justification style and $\theta = 0.968$ for reciprocity style, exceeding the recommended threshold of 0.65 by Passonneau and Carpenter (2014), which indicates that the styles have indeed been incorporated into the outputs by the GPT-3.5 Turbo LLM.

**Discursive features**: Persuasive moves were measured using the Python convokit toolkit, which searches for various lexical features reflective of different categories. These features have been applied to study online persuasion and model politeness and trustworthiness in social media posts (Danescu-Niculescu-Mizil et al. 2013; Niculae et al. 2015).

Regarding Figure 5, GPT-3.5 Turbo-written counter-arguments are typically at par with each other concerning most discursive features. However, they significantly differ

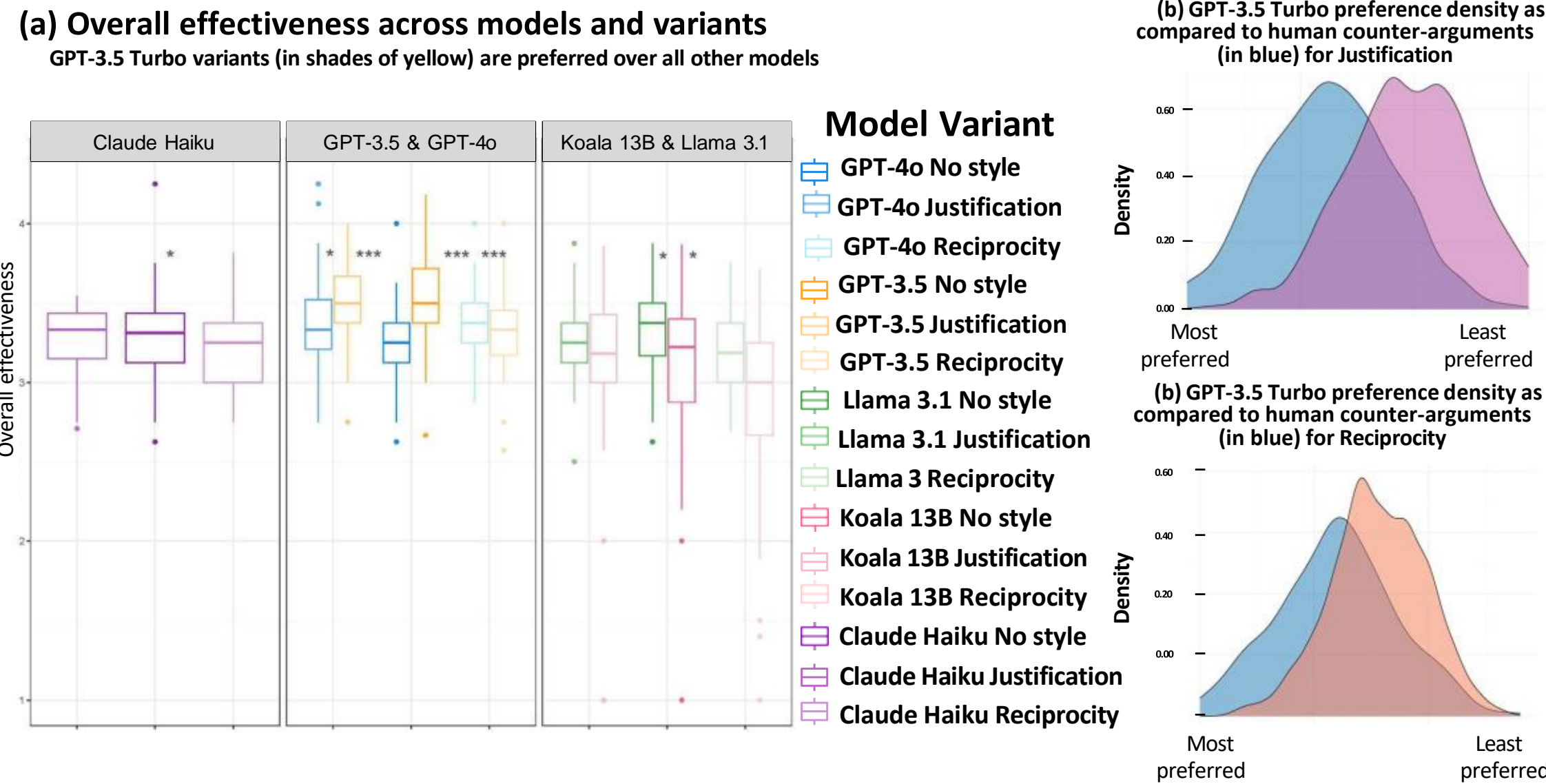

Figure 6: (a) User evaluation of the overall effectiveness of counter-arguments, (b) User preference analysis for human-written (blue) vs. GPT-3.5-written counter-arguments for justification (left) and reciprocity (right) highlights user preferences for reciprocal rather than evidence-based counter-arguments. (1 is most preferred and 4 is least preferred)

($p < 0.001$) from human-written counter-arguments in covering more claims, temporal features, reference to subjects, premises, comparisons, and even politeness. Human-written counter-arguments have fewer claims with greater specificity, offering a more focused and less polite counter-argument.

Overall, the findings suggest that while human-written arguments are richer and more diverse in rhetorical intent, GPT-3.5 Turbo excels in readability metrics, especially in generating more complex texts. This highlights the balance between generating readable content and maintaining rhetorical richness in counter-arguments.

## Rhetorical Insights

We conducted an automated rhetorical analysis to assess the readability and argumentative style of counter-arguments generated by various models and stylistic prompts. Readability metrics (e.g., Flesch-Kincaid, Gunning Fog) revealed that GPT-3.5 Turbo outputs were generally more complex than other models, especially under the No Style and Justification prompts. In contrast, simpler outputs were produced by lightweight models such as Candela. These trends suggest that stylistic prompting increases rhetorical sophistication but also introduces cognitive load.

In terms of rhetorical moves—such as alignment, authority, and persuasion—human-written counter-arguments remain significantly richer and more diverse. They exhibited more agreement/disagreement markers and authority-based expressions (e.g., credentials, experiential claims) than any model-generated outputs. Among LLMs, GPT-3.5 Turbo with Justification and Reciprocity prompts performed better than others, though still lagging behind human benchmarks. Even when re-analyzed using GPT-4o as a rhetorical labeler, model outputs showed sparse use of nuanced rhetorical devices. Full methodological details, prompts, and annotations are available in the appendix and online repository.

## Human Preference Analysis

The first part of the Figure 6 provides insights into the persuasiveness of stylized counter-arguments for the best-performing models relative to the plain outputs, where GPT-3.5 turbo variants have an edge over all the other models in terms of their perceived effectiveness. Furthermore, justification-style counter-arguments have an edge over other variants across the older LLMs, while reciprocity-style arguments appear to improve over no-style counter-arguments in the case of the newer LLMs styles.

The second part of the figure mainly focuses on how GPT-3.5-generated counter-arguments (which scored the highest in the quality and effectiveness scores) fare relative to human-written justification- and reciprocity-style counter-arguments. More specifically, it illustrates that in a comparison of 2000 original posts and counter-arguments sourced from ChangeMyView and the Counterfire corpus, humans find human-written reciprocal-style (Mean preference = 2.24 out of 5; lower score is better) and justification-style counter-arguments (Mean preference = 2.19 out of 5) more preferable to those written by GPT-3.5 turbo (Means 2.93 and 2.56 respectively) (Welch Two Sample t-Test, $p < 0.001$). The low preference for Justification implies that while these directives resulted in comprehensive, evidence-backed counterarguments, they may be less engaging than reciprocal counter-arguments.

Taken together with findings from Figure 5, the findings suggest that the highly focused, specific, and less polite human counter-arguments are more persuasive to humans than GPT-3.5-generated counter-arguments. Although reciprocity counter-arguments are rhetorically richer, they are still perceived to be qualitatively poorer and less effective than the justification-style counter-arguments and even the counter-arguments without style; however, this trend appears to be shifting with the newer LLMs, such as GPT-4o and LLaMA-3.1. The findings suggest an intriguing tradeoff between fact integration and style while generating counter-arguments that inspire future research investigations.

## Error Analysis

A detailed error analysis is provided in the Appendix to better understand how LLMs fare on human evaluation. Furthermore, as some counter-argument generation tasks might favor greater content adherence, we have also reported the counter-arguments scoring high and low on ROUGE-L scores. In general, we observed that examples scoring highly on human quality assessments demonstrate practical and relevant arguments while low-scoring ones often suffer from incompleteness or lack of direct relevance. This was especially the case for counter-arguments from PaLM-2, which scored poorly because of broad statements and lack of direct relevance. Counter-arguments also scored poorly because of their repetitiveness and lack of clarity, especially on counter-arguments on specialized subjects. On the other hand, ROUGE-L F1 scores do not always align with our perception of the substantive quality or originality of a counter-argument.

## Discussion

This study addresses critical research gaps in LLM-generated arguments by examining rhetorical styles—justification and reciprocity—within political discourse. Three years after initiating this work, the problem of balancing persuasive style with evidence-based content remains unsolved. Our findings demonstrate that while modern LLMs can generate factually sound arguments, they struggle with rhetorical nuance essential for genuine persuasion. Furthermore, contemporary metrics like Factscore (Min et al. 2023) capture some nuances of quality, but comprehensive evaluation frameworks accounting for rhetorical effectiveness remain elusive.

Justification emphasizes logical reasoning and evidence-based argumentation, while reciprocity fosters interpersonal engagement with opposing views. Our results show that while LLMs can incorporate these styles to varying degrees, they consistently fail to achieve the rhetorical depth found in human-generated arguments. This highlights a persistent trade-off: high-content arguments often lack the stylistic richness needed for persuasion, while arguments emphasizing reciprocity may sacrifice factual precision. Even advanced models like Claude and GPT-4 produce linguistically complex content but lack the rhetorical diversity of human discourse, revealing fundamental limitations in replicating the nuanced strategies required for effective reasoning and theory of mind. Human evaluations further emphasize this gap, with participants consistently preferring human-generated arguments that balance content with style.

### Contemporary Problems and Future Work

Despite rapid LLM advancement, our intra-model comparisons of counter-argument effectiveness reveal persistent limitations relevant to real-world applications. While models can generate counter-narratives acknowledging opposing viewpoints, their limited rhetorical depth affects their utility in group deliberation (Bose, Reina, and Marshall 2017). The gap between content accuracy and stylistic effectiveness suggests modern LLMs still struggle as autonomous moderators in sensitive contexts (Sinem Tekiroglu, Chung, and Guerini 2020). Additionally, research observes a concerning trend toward sycophancy—models prioritizing perceived user satisfaction over genuine reasoning (Rimsky et al. 2024)—further eroding persuasive integrity.

Recent work on AI-mediated deliberation (Tessler et al. 2024) demonstrated that an LLM-based “Habermas Machine” can help diverse groups converge on shared perspectives, outperforming human mediators in clarity, informativeness, and fairness. In contrast, our findings suggest that while LLMs are capable of producing stylized and evidence-based counter-arguments that humans prefer over generic outputs, they still fall short of human standards in rhetorical richness and persuasive strength—especially when evaluated at the level of individual argumentative moves. This divergence highlights a fundamental distinction in goals: whereas Tessler et al. prioritize convergence, we emphasize the value of discursive diversity and rhetorical nuance in deliberative quality. Nevertheless, their findings offers an interesting potential application for our work.

Future work must address persistent challenges in LLM reasoning and theory of mind capabilities (Tan et al. 2025). Development of dynamic models capable of tailoring arguments to conversational partners’ stylistic preferences could enhance persuasive power (Salvi et al. 2024), potentially mitigating sycophantic tendencies (Rimsky et al. 2024). Since persuasion typically evolves across multiple exchanges, future research should explore the relationship between argument style and persuasive effect over extended interactions (Jaidka, Ahuja, and Ng 2024). Models designed for persuasive argumentation would benefit from temporal sequence training to simulate evolving discourse. Advanced techniques like model unlearning could help eliminate undesirable style biases (Yuan et al. 2025), while maintaining the critical balance between evidence-based reasoning and persuasive rhetoric that continues to elude even the most sophisticated contemporary systems (Li and Goyal 2025). We have discussed the limitations of our work in the appendix.

### Ethical Considerations in the Study of LLM Persuasiveness

This section was added post-acceptance, while preparing the camera-ready submission. In light of recent revelations of research malpractice in LLM persuasion experiments (CMV Mod Team 2025), we felt compelled to justify this line of research, but also defend and situate our work at this juncture. There have been significant past controversies where field experiments on social media raised substantial public concerns about human subjects research, most notably Face-book's 2014 "emotional contagion" study that manipulated the news feeds of nearly 700,000 users without explicit consent (Selinger and Hartzog 2016; Jouhki et al. 2016). While survey experiments like ours avoid direct deception of human participants, they still require critical examination of potential downstream harms from the dissemination of techniques enhancing LLM persuasiveness. Some ethical failures sometimes emerge from inexperienced junior researchers navigating complex research environments without adequate oversight or training in research ethics specific to AI (Metcalf et al. 2021). Regardless of origin, all authors share responsibility for research misconduct, though career consequences often fall disproportionately on junior team members, creating an inequitable distribution of reputational damage.

Another critical consideration is the researcher's extensive degrees of freedom in LLM-based studies. With countless parameter combinations, prompt engineering approaches, and selective reporting opportunities, the risk of p-hacking and publication bias is substantially heightened (Bakker et al. 2021). Particularly troubling is the methodological challenge of "prompt hacking"— potentially circumventing built-in ethical safeguards or approved research protocols through engineered prompts (Perez et al. 2022). This practice effectively bypasses guardrails established through institutional oversight mechanisms. Finally, the use of deception in field experiments that use LLMs for persuasion on public social platforms like Reddit exemplify concerning practices, including explicit deception toward both community members and platform moderators, as well as deceptive research practices when prompting LLMs. Such practices fundamentally violate research integrity principles and exploit online communities developed specifically for good-faith communication, contravening ethical standards established for research on consenting human subjects research (National Commission for the Protection of Human Subjects of Biomedical and Behavioral Research 1979) and establishing concerning precedents. Post-treatment biases, such as discarding data from deleted posts without accounting for potential selection effects, severely limit experimental replicability and demonstrate how methodological choices can undermine validity while artificially amplifying persuasive effects (Montgomery, Nyhan, and Torres 2018). These issues underscore the need for more conscientious research practices and ethical reflection before undertaking projects involving online communities. For instance, exemplary scholars have previously partnered with r/ChangeMyView moderators to conduct impactful research on discussion norms and online participation (Matias 2019).

Our findings on the rhetorical capabilities of LLMs demand careful consideration of broader societal implications. The capacity to generate persuasive yet potentially unfounded arguments that incorporate justification and reciprocity styles raises the specter of automated misinformation campaigns with enhanced rhetorical sophistication (Weidinger et al. 2022). Yet, the persistent gap between rhetorical effectiveness and factual accuracy that we've identified suggests that advancing this research responsibly requires not just technical refinement but collaborative engagement between AI researchers, ethicists, and communication scholars to establish norms that protect deliberative integrity in increasingly AI-mediated discourse environments (Bender et al. 2021). Rather than simply documenting these capabilities, responsible research must actively develop countermeasures and detection methods while advocating for transparent LLM deployment practices (Gebru et al. 2021; Liu et al. 2024). The field requires robust ethical guidelines specifically addressing LLM research. These should include: mandatory transparency about LLM involvement in content generation; prohibition of deceptive practices in prompt engineering; comprehensive pre-registration of methodological choices; and integrated ethical review beyond traditional human subjects frameworks to account for potential societal impacts.

## Acknowledgements

This research was supported by the Ministry of Education, Singapore, through its MOE AcRF Tier 3 Grant (MOE-MOET32022-0001) and the Tier 1 programme (WBS A-8000231-01-00). We acknowledge the support of the Centre for Trusted Internet and Community and the Institute of Data Science at the National University of Singapore. We also extend our gratitude to the members of the NUS Communications and New Media journal club for their valuable feedback.

## Paper Checklist

1. For most authors...
   (a) Would answering this research question advance science without violating social contracts, such as violating privacy norms, perpetuating unfair profiling, exacerbating the socio-economic divide, or implying disrespect to societies or cultures? Yes
   (b) Do your main claims in the abstract and introduction accurately reflect the paper's contributions and scope? Yes
   (c) Do you clarify how the proposed methodological approach is appropriate for the claims made? Yes, in the Method and Results section
   (d) Do you clarify what are possible artifacts in the data used, given population-specific distributions? Yes, in the Limitations section
   (e) Did you describe the limitations of your work? Yes, in the Limitations section
   (f) Did you discuss any potential negative societal impacts of your work? Yes, in the Ethics statement
   (g) Did you discuss any potential misuse of your work? Yes, in the Ethics statement
   (h) Did you describe steps taken to prevent or mitigate potential negative outcomes of the research, such as data and model documentation, data anonymization, responsible release, access control, and the reproducibility of findings? Yes, in the Limitations section
   (i) Have you read the ethics review guidelines and ensured that your paper conforms to them? Yes
2. Additionally, if your study involves hypotheses testing...
   (a) Did you clearly state the assumptions underlying all theoretical results? Not applicable
   (b) Have you provided justifications for all theoretical results? Not applicable
   (c) Did you discuss competing hypotheses or theories that might challenge or complement your theoretical results? Not applicable
   (d) Have you considered alternative mechanisms or explanations that might account for the same outcomes observed in your study? Not applicable
   (e) Did you address potential biases or limitations in your theoretical framework? Not applicable
   (f) Have you related your theoretical results to the existing literature in social science? Not applicable
   (g) Did you discuss the implications of your theoretical results for policy, practice, or further research in the social science domain? Not applicable
3. Additionally, if you are including theoretical proofs...
   (a) Did you state the full set of assumptions of all theoretical results? Not applicable
   (b) Did you include complete proofs of all theoretical results? Not applicable
4. Additionally, if you ran machine learning experiments...
   (a) Did you include the code, data, and instructions needed to reproduce the main experimental results (either in the supplemental material or as a URL)? Yes, in the supplementary materials and the online repository
   (b) Did you specify all the training details (e.g., data splits, hyperparameters, how they were chosen)? Yes, in the supplementary materials and the online repository
   (c) Did you report error bars (e.g., with respect to the random seed after running experiments multiple times)? Yes, in all the figures
   (d) Did you include the total amount of compute and the type of resources used (e.g., type of GPUs, internal cluster, or cloud provider)? Yes, in the supplementary materials and the online repository
   (e) Do you justify how the proposed evaluation is sufficient and appropriate to the claims made? Yes, in the Methods and Results and the Discussion
   (f) Do you discuss what is "the cost" of misclassification and fault (in)tolerance? Yes, in the Limitations section
5. Additionally, if you are using existing assets (e.g., code, data, models) or curating/releasing new assets, **without compromising anonymity**...
   (a) If your work uses existing assets, did you cite the creators? Yes
   (b) Did you mention the license of the assets? No
   (c) Did you include any new assets in the supplemental material or as a URL? Yes, in the supplementary materials and the online repository
   (d) Did you discuss whether and how consent was obtained from people whose data you're using/curating? Not applicable, as we are not using data about people
   (e) Did you discuss whether the data you are using/curating contains personally identifiable information or offensive content? Yes, in the Limitations section
   (f) If you are curating or releasing new datasets, did you discuss how you intend to make your datasets FAIR? Yes, in the Ethics Statement
   (g) If you are curating or releasing new datasets, did you create a Datasheet for the Dataset? Yes, it is included in the supplementary materials.
6. Additionally, if you used crowdsourcing or conducted research with human subjects, **without compromising anonymity**...
   (a) Did you include the full text of instructions given to participants and screenshots? Yes
   (b) Did you describe any potential participant risks, with mentions of Institutional Review Board (IRB) approvals? No
   (c) Did you include the estimated hourly wage paid to participants and the total amount spent on participant compensation? Yes, in the supplementary materials
   (d) Did you discuss how data is stored, shared, and deidentified? Not applicable, as the data does not include any personally identifiable information.

## Ethics Statement

The dataset comprises public threads from the subreddit. There was no personal data used. Automatic measurements are privy to model accuracy, which are not readily available for domain-specific applications. The prompts developed in this work may only generalize to some contexts. We observed that including snippets from news articles or Wikipedia can lead us to inadvertently quote individuals in the public eye as part of the arguments. For instance, some evidence includes the names of experts, politicians, and the heads of state if they were included in a relevant article. This information must be reviewed and redacted before a public rollout or implementation based on the Counterfire corpus. Furthermore, given that the Counterfire corpus is intended for auditing, it would be dangerous to finetune models on this dataset without masking or verifying its factual references or assumptions.

The following two ethical considerations concern the use of our framework and its implications. First, while our work helps to develop a deeper understanding of the principles of argumentation, with applications to understanding persuasion and trustworthiness. However, modeling these negotiation strategies with generative models may have implications for vulnerable audiences; for instance, models finetuned on the labeled dataset could work to gain someone's trust with malicious intent or mislead them in some manner. Second, the swift updates to LLMs limits the replicability of our results, where it was necessary to rerun our methods on the state of the art every few months. Nevertheless, because our analysis focuses on intra-model comparisons of counter-argument effectiveness, the set of findings can easily convey our main message regarding the need to bridge the work on style and argumentation for real-world application.

## Appendix

## Datasheet for the Counterfire Dataset

### Motivation

**For what purpose was the dataset created?**
The Counterfire dataset was created to study stylized counter-argument generation in political argumentation, with a focus on evaluating the balance between evidence integration and rhetorical style in counter-arguments. It was designed to fill a gap in understanding how stylistic variations impact human preferences and persuasiveness in debates.

**Who created the dataset?**
The authors of this submission. The names are not included to respect ICWSM's anonymity policy.

**Who funded the creation of the dataset?**
The research was funded by the authors, with support from affiliated research institutions. There was no external grant specifically associated with this dataset creation. The names are not included to respect ICWSM's anonymity policy.

### Composition

**What do the instances that comprise the dataset represent?**
The instances represent counter-arguments generated in response to original posts from the ChangeMyView subreddit. Each instance includes the original post, retrieved evidence from sources (e.g., Wikipedia, news articles), and counter-arguments generated by different models such as GPT-3.5 Turbo and Koala-13B.

**How many instances are there in total?**
The dataset consists of 38,000 counter-arguments generated from 2,000 original posts, across three distinct styles: plain, reciprocity, and justification.

**Does the dataset contain all possible instances?**
No, the dataset is a sample of ChangeMyView posts. It was curated to include diverse topics and argument types commonly found in online political discussions.

**What data does each instance consist of?**
Each instance consists of the original post, retrieved evidence, and a generated counter-argument, along with metadata about the style and model used.

**Is there a label or target associated with each instance?**
Yes, the dataset includes human preference labels for quality and persuasiveness, as well as labels indicating adherence to rhetorical styles (justification, reciprocity).

**Is any information missing from individual instances?**
No information is intentionally missing from individual instances.

**Are relationships between individual instances made explicit?**
Connections between original posts and counter-arguments can be made.

**Are there recommended data splits?**
No, none

**Are there any errors, sources of noise, or redundancies in the dataset?**
We expect that automatic generation of counter-arguments could introduce hallucination or other semantic inconsistencies.

**Is the dataset self-contained, or does it link to external resources?**
The dataset is self-contained, but it relies on a static dataset of external resources for retrieving evidence, such as Wikipedia and news websites. These external resources may evolve over time.

**Does the dataset contain data that might be considered confidential?**
No, the dataset only contains public posts and counter-arguments. No confidential data is included.

**Does the dataset contain data that might be offensive, insulting, or threatening?**
Some original posts may contain content related to sensitive political discussions, but such content is typical of public forums and is not considered overtly offensive or harmful.

### Collection Process

**How was the data associated with each instance acquired?**
The original dataset of original posts, human-written counter-arguments, and evidence selected from the domain knowledge relevant to the posts was made available by Hua, Hu, and Wang (2019).

**What mechanisms were used to collect the data?**
Counter-arguments for these inputs (excluding the human counter-argument) were generated by various language models based on different prompts.

**If the dataset is a sample, what was the sampling strategy?**
We used the entirety of the dataset from Hua, Hu, and Wang (2019), setting aside 2000 instances for counter-argument generation while using 70,000 instances for model fine-tuning on human-generated counter-arguments.

**Who was involved in the data collection process?**
The authors and their research assistants were involved in data collection.

**Over what timeframe was the data collected?**
Data collection took place over a period of eighteen months, ensuring a variety of LLMs from that period were included.

**Were any ethical review processes conducted?**
No formal ethical review was conducted, as the data was collected from publicly available sources.

### Preprocessing/Cleaning/Labeling

**Was any preprocessing/cleaning/labeling of the data done?**
Yes, preprocessing included cleaning the original posts to remove unnecessary metadata.

**Was the raw data saved in addition to the preprocessed data?**
Yes.

**Is the software that was used to preprocess the data available?**
No specific software tools were used for preprocessing that are available publicly, though the code used for retrieval and generation is shared in the project repository.

### Uses

**Has the dataset been used for any tasks already?**
No, not at the moment.

**Is there a repository linking to papers or systems that use the dataset?**
No, not at the moment, but in the future, links to relevant papers and systems will be provided in the project's online repository.

**What (other) tasks could the dataset be used for?**
The dataset can be used for:

- Studying counter-argument generation.
- Analyzing human preferences in online debates.
- Investigating rhetorical strategies in political communication.

**Is there anything about the composition of the dataset that might impact future uses?**
The dataset reflects the dynamics of the ChangeMyView subreddit, and some biases inherent to the community may be present. These have been discussed in the Discussion section and the Ethics Statement.

**Are there tasks for which the dataset should not be used?**
The dataset is not suitable for tasks requiring data related to private or sensitive topics, as it contains only publicly available posts. It has also not been vetted for informational accuracy. It should only be used for the purpose of understanding and modeling stylized arguments.

### Distribution

**How will the dataset be distributed?**
The dataset is available at https://github.com/Preetika764/Style_control/.

**What license is the dataset distributed under?**
The dataset is distributed under a Creative Commons Attribution 4.0 International (CC BY 4.0) license.

### Maintenance

**Who will be supporting/maintaining the dataset?**
The dataset will be maintained by the authors.

**Will the dataset be updated?**
There are no current plans to update the dataset. Any updates will be announced on the distribution page.

## Hyperparameter settings

The Bitsandbytes wrapper was used for quantization. LoRa was applied to the base model after loading in 4 bits. The following were the specific LoRA hyperparameters:

- rank of update matrices = 8
- dropout = 0.05
- target modules = q and v attention matrices
- LoRA scaling factor = 32
- all params = 6678533120
- trainable params = 6553600
- trainable % = 0.0981

The following were the fine-tuning hyperparameters:

- per_device_train_batch = 1
- learning rate = 0.0002
- optimizer = Paged Adam 8bit optimizer

Figure 7 reports the training loss plots for GPT-3.5-turbo and Koala-13B fine-tuning.

The configuration parameters when we prompted GPT-3.5 turbo, GPT-3.5-finetuned and GPT4o for text generation were the default settings: N-epochs: 4, learning-rate-multiplier: 0.1.

The configuration parameters for generating text with Koala-13B and Koala-13B-finetuned were: max_new_tokens: 120, temperature: 1, topK: 50, topP: 1.

Finally, the configuration parameters for PaLM-2 were: temperature: 0.8, maxOutputTokens: 256; topK: 40; topP: 0.95.

The configuration parameters for LLaMa-3.1 were: temperature: 1, top_p: 0.9, top_k: 50.

The configuration parameters for generating text with Claude Haiku were: temperature: 1, top_p: 1, top_k: 0.

### Dataset Definitions

In this paper, we refer to several datasets using shorthand notations for brevity:

1. **Candela** — The *Candela* dataset (Hua, Hu, and Wang 2019) contains English-language posts and corresponding counter-arguments collected from the r/ChangeMyView subreddit.

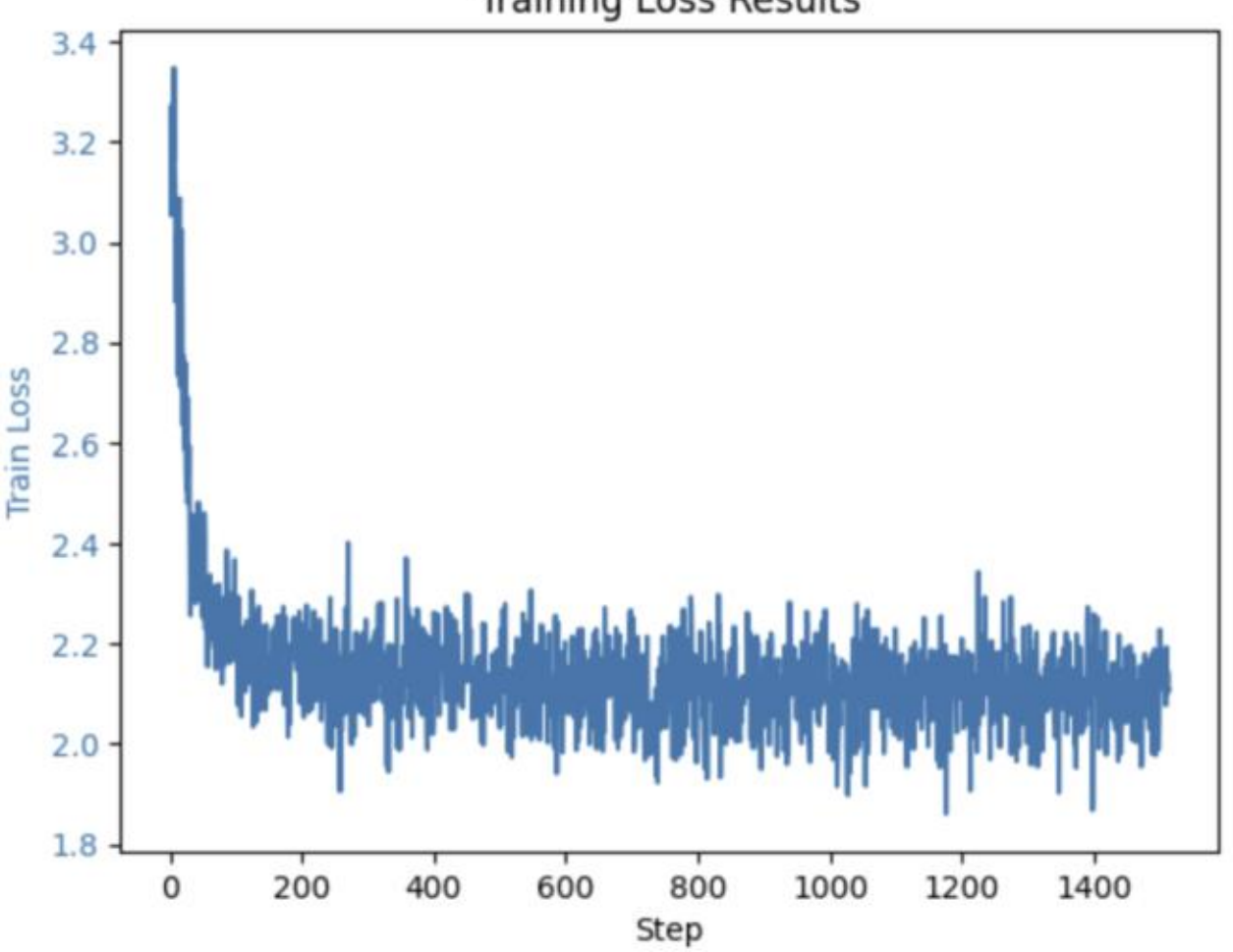

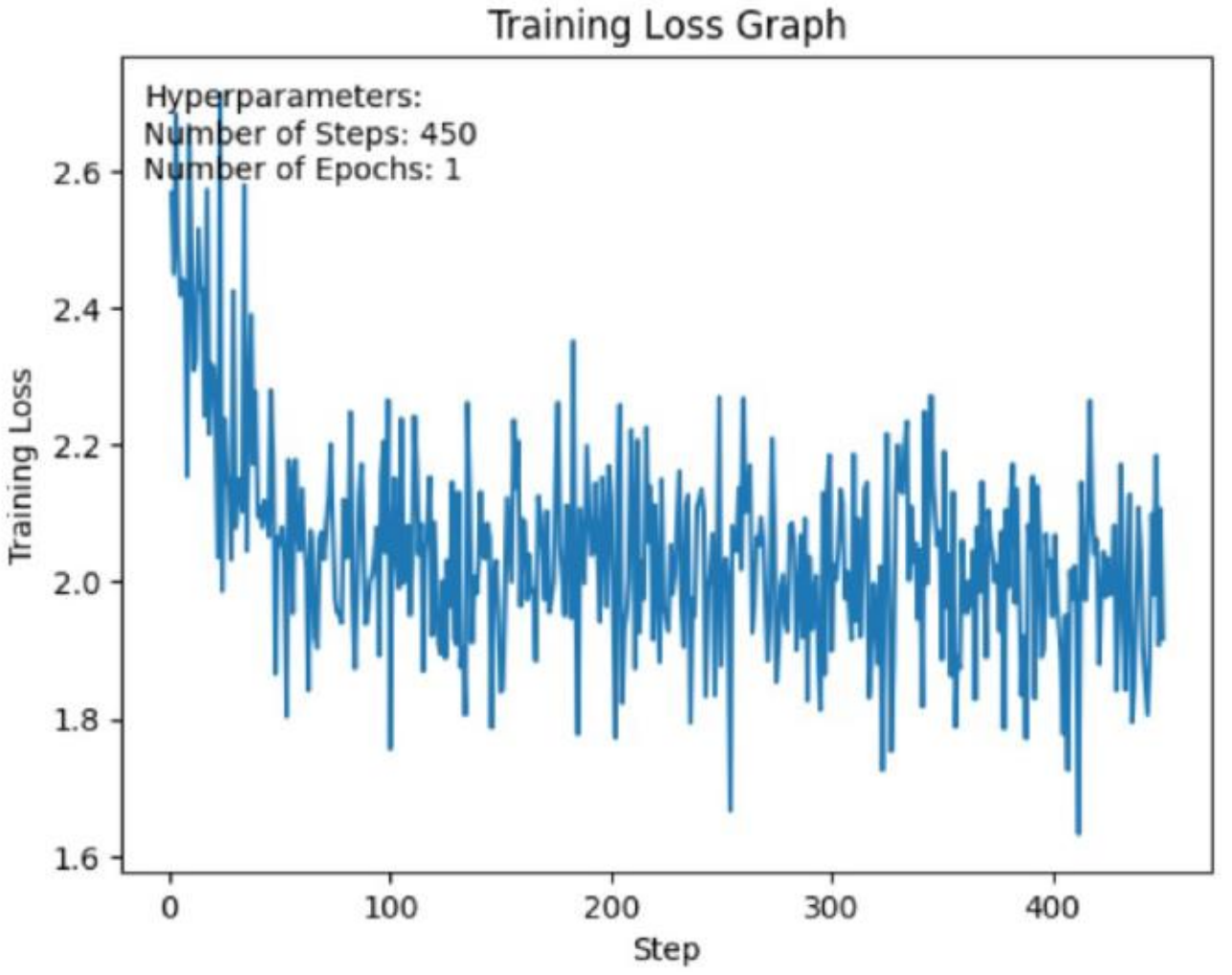

Figure 7: Fine-tuning training loss plots for (a) GPT-3.5 turbo and (b) Koala-13B

2. **ChangeMyView (CMV)** — The *CMV* dataset refers to content from the r/ChangeMyView subreddit, a Reddit community where users share their opinions and invite others to challenge their views through reasoned discussion. This subreddit serves as the source for both the Candela and Counterfire datasets.
3. **Counterfire** — We introduce *Counterfire*, a new dataset of 38,000 counter-arguments by generating responses for 2000 inputs from the Candela dataset. Specifically, we generate responses in three distinct styles: *reciprocity*, *justification*, and *no style*, facilitating the study of style-controlled argument generation.

For clarity, the shorthand notations (*Candela*, *CMV*, and *Counterfire*) will be used throughout the paper.

## Argument Style and Quality Annotation

A random sample of 100 corresponding counter-arguments generated for the same inputs by each of the LLM variants was included in an Amazon Mechanical Task to get eight annotations per argument on the quality of the text and five annotations per argument on the discussion facet labels of justification and reciprocity (in a different HIT). Amazon Mechanical Turkers who had completed at least 10,000 HITs, were residents of the USA, and had an approval rating of 95% or above were eligible to participate in the task. They were paid $0.10 per HIT. Their demographic information was not collected.

The eight annotation labels per facet per argument thus obtained were used to create Figure 4. The evaluation scores are included in the online repository. Below, we report the codebook applied to evaluate the different facets of argument quality.

Figure 8 and Figure 9 illustrate some outputs from the Koala-13B and PaLM-2 models. We observed that PaLM-2 outputs were inconsistent in adhering to the instructions for generating a counter-argument. In about 10% of the cases, it generated an argument in support of the opinion instead.

**Input statement:** employers should not be allowed to fire employees over their political views or humorous statements made outside of work . i recently saw a case on twitter where a man tweeted a transphobic joke where he posted a screenshot of a news article that said the (...)
**Evidence:** ["i 'm not sure what you mean by "censorship". i don't think it's a good idea to do this. it's not about encouraging people to express their views.","if it is the latter, you need to address the issue forthrightly with your supervisor, and it may even be appropriate for you to report the matter to your organization's hr or compliance officer", "if we're really going to change the dynamic in the workplace, we need men to stop being enablers and bystanders and come forward"...]

**Candela output:** here 's a real life example of what would transpire if employers could not fire people for their political views .(...)

**Koala output with no style:** The argument that employers should not be allowed to fire employees over their political views or humorous statements made outside of work is not only unethical but also sets a bad precedent for society. The argument is based on the idea that (...)

**Koala output with reciprocity:** As a socialist, I believe that workers have a right to a separate personal life to their professional life and that no non-criminal activity done while not on the clock and unrelated to their job should (...)

**Koala output with justification:** The argument that employers should not be allowed to fire employees over their political views or humorous statements made outside of work is a controversial one. While it is true that employers have the right to hire and fire employees based on their political views, (...)

Figure 8: An example input statement and the generated counter-arguments for the CMV dataset from Koala-13B.

**Instructions for Style Validation** Annotators (Amazon Mechanical Turk workers who were residents of the United States with at least a 96% approval rate who had at least 5000 approved HITs) were provided with the title of the original post followed by a counter-argument. The following was the task description for the style validation task:
In this job, you will be presented with comments made in reply to a post on Reddit, a popular discussion forum worldwide. The topic of the discussion is in the "Title." It is a conversation that the OP started in a community called Change My View. They post there because they want to discuss the topic with people who have a different point of view, maybe to learn something new or to change their mind about it.

**Input statement:** cmv: privacy is a fundamental human right and we should stop using the products of corporations that violate it. we should start questioning whether the business models of these huge data-gathering corporations are ethical and benefit society. do we want to live in a world where a few handful of billionaires control the information about our personal lives? (...)
**Evidence:** ["(...) most consumers don't read privacy policies , which only makes google 's reluctance to make it available off of their home page all the more open to questioning. (...)","i can't say that everything i look at, i own, and expect to be taken seriously. if it comes to that, most people would never install much of the software out there ( or use their isp provider or wireless company ) if they really read the accompanying tos or eula – for in many cases they effectively strip the user of any rights whatsoever (...)", "google sells advertising – telling other people what they do not want to know and explain that what they consider lesser to be "better" (...)]

**Candela output:** i don't think it's a form of censorship, it's just a collection of information. you can't just say "you can't have privacy" (...). If you don't want to publish information, you're not going to be able to do what you want with the information you're supposed to do (...)

**PaLM2 output with no style:** While privacy is important, the benefits of using technology often outweigh the risks. These products and services enhance our lives, making them more convenient, efficient, and enjoyable. Moreover, companies are increasingly implementing privacy-protecting measures. Many now offer users more control over their data, allowing them to opt out of data collection or delete their (...)

**PaLM2 output with reciprocity:** While privacy is important, it is not a fundamental human right. There are many other rights that are more important, such as the right to life, the right to freedom of speech, and the right to a fair trial. Additionally, there are some cases in which it is necessary to sacrifice privacy in order to achieve other important goals, such as national security or (...)

**PaLM2 output with justification:** While it's true that privacy is a fundamental human right, the benefits of data collection and analysis far outweigh the risks. **Evidence 1:** Open data initiatives have empowered researchers and policymakers to address critical societal challenges, such as climate change, public health, and economic inequality. **Evidence 2:** Data-driven journalism has played a vital role (...)

Figure 9: An example input statement and the generated counter-arguments for the CMV dataset from PaLM-2.

You are reading a comment by a Commenter on the post. They are trying to persuade the OP. Review the text of the comment and help us by answering a few yes/no questions about it. Each HIT takes about 30 seconds.

1. Read the comment.
2. Determine which categories best describe the comment.
   - **RECIPROCITY Description:**
     - YES: Whether this comment asks questions or tries to get a response from someone about their opinions or information sources. Examples:
       - *Could you please share copies or provide relevant links to the information?*
       - *How did the naming of Chad in the travel ban impact Niger?*
       - *What's the reason behind your sponsorship of legislation to halt the Russia investigation?*
       - *When you say "Would have preferred," it implies you're somewhat okay with the current situation but would have liked another outcome. Is this your genuine sentiment? Did someone influence your opinion?*
       - *The tax bill seems to require more than just minor adjustments. It appears to need a complete overhaul. Why not just reject it?*
       - *It's evident that Trey Gowdy speaks assertively, but when will we see him take decisive actions to match his words?*
       - *What criteria determine a credible source? There are politicians who base their decisions on questionable sources, so how can the legitimacy of such sources be legally challenged?*
       - *Considering the original intent of the minimum wage was to ensure a living wage, as stated by FDR, how has this vision evolved over time?*
     - NO: This comment does not ask a genuine question or asks rhetorical questions.
   - **JUSTIFICATION Description:**
     - YES: Personal: Whether this comment contains personal feelings or experiences. Examples:
       - *Corporate Democrats, be aware that we're watching closely. You're on notice.*
       - *Senator [name] from the Republican party stated, "We all recognize that [name] is not up to the mark."*
       - *It seems like [name] has been given a blank check. Their credibility is questionable at this point.*
       - *It's essential to stay informed and make our voices heard. If our representatives don't shape up, we'll vote them out.*
     - YES: Fact-based: Whether this comment contains facts, links, or evidence from other sources. Examples:
     - NO: This comment does not offer a justification.

**Instructions for Quality Evaluation** Annotators (Amazon Mechanical Turk workers who were residents of the United States with at least a 96% approval rate who had at least 5000 approved hits) were provided with the title of the original post, followed by a counter-argument. The following were the instructions for the task: These are arguments posted on Reddit in response to an original argument.
Please classify them according to various facets.

**Grammar:**

- Poor: The statement contains many grammatical errors and is difficult to understand.
- Fair: The statement contains some grammatical errors that may affect clarity.
- Good: The statement is generally grammatically correct but may contain occasional errors.
- Excellent: The statement is well-written and largely free of grammatical errors.
- Flawless: The statement is flawless in its grammar and syntax.

**Content richness:**

- Poor: The argument is extremely shallow and lacks substance.
- Fair: The argument is somewhat lacking in substance and may be overly simplistic.
- Good: The argument has some substance, but may lack depth or nuance.
- Excellent: The argument is rich and detailed, with plenty of supporting evidence and nuanced arguments.
- Flawless: The argument is extremely rich and detailed, with complex arguments and a wealth of supporting evidence.

**Logic and reasoning:**

- Poor: The argument is illogical and poorly reasoned.

- Fair: The argument is somewhat illogical and poorly reasoned.
- Good: The argument is neither well nor poorly reasoned, and has some logical flaws.
- Excellent: The argument is quite logical and well-reasoned.
- Flawless: The argument is very logical and flawlessly reasoned.

**Overall effectiveness:**

- Poor: The argument is very weak and fails to convince me.
- Fair: The argument is somewhat weak and unconvincing.
- Good: The argument is neither strong nor weak, and is somewhat convincing.
- Excellent: The argument is quite strong and convincing.
- Flawless: The argument is very strong and completely convincing.

## Instructions for User Preference Analysis

The original post was presented to each survey respondent, followed by four counter-arguments: the human-written argument from the Candela dataset, and three variants from the GPT-3.5-turbo. The survey was launched on Amazon Mechanical Turk to residents of the United States with at least a 96% approval rate who had at least 5000 approved hits. The median age was 34.5 years. 691 (36.7%) were female, and 854 (45.4%) were male, while 74 (3.9%) identified as non-binary or third gender. The remaining respondents did not share their age nor gender.

The following was the description of the task:
In this job, you will be presented with various counter-arguments posted in the ChangeMyView subreddit. In ChangeMyView, users present a viewpoint, and others respond with counter-arguments to challenge or change the original viewpoint. Your role is to read these counter-arguments and assess their effectiveness in persuading against the Original Post. Consider the logic, evidence, and clarity of each argument in your evaluation. Each HIT will take approximately 2-3 minutes, depending on the length and complexity of the arguments. Pay attention to the strength of the reasoning and the use of evidence in each counter-argument.

The following were the step-by-step instructions:

- These are counter-arguments posted in response to an “Original Post” within a Reddit community called ChangeMyView.
- Each counter-argument is an attempt to persuade people against the viewpoint presented in the Original Post.
- Your task is to evaluate and order these counter-arguments based on their persuasiveness.
- According to your preference, please state whether you agree with the opinion in the original post.
- Next, at least once for this batch of HITs, please share your age and gender. These questions are optional.
- Finally, according to your preference, please rank the arguments, with the most persuasive argument as #1.

# Additional Results

## Inter-Annotator Reliability

Based on our choice of style prompts and the related prior work (Goyal, Li, and Durrett 2022; Wachsmuth et al. 2017), our evaluation focused on **content, grammar, logic, and overall effectiveness**. The inter-annotator agreement statistics are reported in Table 4 and indicate that the annotation quality is reliable ($\theta$ >0.65).

| **Human annotation of argument quality** | |
|---|---|
| | **$\theta$ (Inter-annotator accuracy $\theta$)** |
| **Content** | 0.8395 |
| **Grammar** | 0.8831 |
| **Logic** | 0.8891 |
| **Overall effectiveness** | 0.8951 |

Table 4: Inter-annotator reliability statistics. $\theta$ is the average annotator accuracy across true-positives and negatives (Passonneau and Carpenter 2014).

## Human evaluation

Table 5 reports the human evaluation scores for all the models and indicates Bonferroni-corrected statistical significance ($p<0.05$) where the quality of one variant is significantly better than other outputs.

## Rhetorical Insights

We conducted an automatic rhetorical analysis to compare the generated counter-arguments along many dimensions of readability, rhetorical intent, and discourse.

**Readability Metrics:** We used the textstat package to calculate various readability metrics, such as Flesch-Kincaid grade, Flesch Reading Ease, Gunning Fog index, and Smog index. In Table 6, we provide a comprehensive evaluation of readability and rhetorical intent across different models and prompting strategies. The readability metrics indicate that among GPT-3.5 Turbo outputs, No Style has the highest score (12.81), indicating higher complexity, followed by Justification (12.75) and Reciprocity (11.79). Similar trends are observed across other readability metrics, with GPT-3.5 Turbo No Style demonstrating the most complexity in terms of Flesch Reading Ease, Gunning Fog, and Smog Index. In comparison, outputs from Candela had the lowest complexity (6.40).

**Rhetorical Intent:** We characterized the generated counter-arguments according to the frequency counts of the phrases related to rhetorical moves for argument alignment, authority, and persuasion. The Alignment and Authority in Wikipedia Discussions (AAWD) corpus (Bender et al. 2011) provided a basis for counter-argument analysis, with annotated phrases indicating agreement or disagreement. Authority moves express credibility, while persuasive moves include features like politeness, contingency, expansion, claims, and premises.

| Model Variant | Content | Grammar | Logic | Overall |
|---|---|---|---|---|
| Candela | 2.99 (0.34) | 2.91 (0.33) | 3.02 (0.34) | 2.91 (0.33) |
| GPT 3.5 Turbo - Justification | 3.47 (0.29)*** | 3.7 (0.28)*** | 3.47 (0.3)*** | 3.52 (0.27)*** |
| GPT 3.5 Turbo - No style | 3.53 (0.28)*** | 3.62 (0.34)*** | 3.42 (0.3)*** | 3.53 (0.27)* |
| GPT 3.5 Turbo - Reciprocity | 3.3 (0.28) | 3.51 (0.34) | 3.29 (0.34) | 3.34 (0.28) |
| GPT 3.5 Turbo Finetuned - Justification | 3.2 (0.3) | 3.19 (0.27) | 3.22 (0.35) | 3.22 (0.33) |
| GPT 3.5 Turbo Finetuned - No style | 3.18 (0.32) | 3.22 (0.27) | 3.26 (0.29) | 3.19 (0.3) |
| GPT 3.5 Turbo Finetuned - Reciprocity | 3.24 (0.34) | 3.22 (0.29) | 3.25 (0.38) | 3.25 (0.37) |
| GPT 4o - Justification | 3.38 (0.29) | 3.4 (0.24) | 3.4(0.26) | 3.39 (0.31)* |
| GPT 4o - No style | 3.29 (0.26)* | 3.37 (0.28)* | 3.31 (0.28)* | 3.24 (0.25)** |
| GPT 4o - Reciprocity | 3.41 (0.23) | 3.39 (0.25) | 3.41 (0.22) | 3.37 (0.21) |
| Koala - Justification | 3.03 (0.61) | 3.28 (0.6) | 2.96 (0.6) | 3.06 (0.62)* |
| Koala - No style | 2.97 (0.62) | 3.29 (0.63) | 2.96 (0.65) | 3.04 (0.62) |
| Koala - Reciprocity | 2.8 (0.69) | 3.1 (0.62) | 2.75 (0.65) | 2.81 (0.66) |
| Koala Finetuned - Justification | 2.19 (0.33) | 2.38 (0.35) | 2.18 (0.32) | 2.21 (0.32) |
| Koala Finetuned - No style | 2.17 (0.39) | 2.37 (0.37) | 2.11 (0.34) | 2.18 (0.39) |
| Koala Finetuned - Reciprocity | 2.09 (0.36) | 2.38 (0.35) | 2.11 (0.37) | 2.15 (0.38) |
| Llama 3 - Justification | 3.3 (0.23) | 3.4 (0.23)* | 3.28 (0.3) | 3.25 (0.26) |
| Llama 3 - No style | 3.31 (0.25) | 3.27 (0.28)* | 3.33 (0.33) | 3.33 (0.27)* |
| Llama 3 - Reciprocity | 3.32 (0.29) | 3.41 (0.24) | 3.21 (0.31) | 3.19 (0.26) |
| PALM - Justification | 3.27 (0.25) | 3.31 (0.26) | 3.3 (0.3) | 3.31 (0.29) |
| PALM - No style | 3.27 (0.34) | 3.3 (0.26) | 3.3 (0.32) | 3.28 (0.32) |
| PALM - Reciprocity | 3.31 (0.29) | 3.32 (0.23) | 3.29 (0.24) | 3.29 (0.29) |
| Claude - Justification | 3.33 (0.25) | 3.37 (0.24) | 3.3 (0.3) | 3.29 (0.2) |
| Claude - No style | 3.38 (0.3) | 3.38 (0.23) | 3.36 (0.25)* | 3.29 (0.27) |
| Claude - Reciprocity | 3.29 (0.22) | 3.34 (0.27) | 3.24 (0.26) | 3.21 (0.26) |

Table 5: Results from the human evaluation on various dimensions: content, grammar, logic, and overall effectiveness scores. The values represent the mean (SD) for each metric. Asterisks (*) denote that the model variant significantly outperformed others in the same group based on pairwise comparisons using a Bonferroni-corrected statistical significance (*** :$p < 0.001$; ** :$p < 0.01$; * :$p < 0.05$).

- **Alignment Moves:** Human-written counter-arguments contain the most positive and negative moves (12 each), highlighting rich argumentative content. These moves are examples of social acts involving agreement or refutation in argumentation. In contrast, GPT-3.5 Turbo variants have fewer alignment moves. The Justification style includes more positive (4) and negative (4) moves than Reciprocity.
- **Authority Moves:** Authority moves are markers of social expectations, credentials, experiential claims, forum claims, and external claims. Human-written arguments lead across all categories. Specifically, the Reddit counter-arguments contain 12 positive and negative alignment moves, showing explicit agreement and positive alignment (e.g., praise thinking), as well as opposing alignment (e.g., criticizing or doubting). GPT-3.5 Turbo Reciprocity includes a notable number of experiential (6) and forum (4) moves, while Justification leads in external (4) moves.

Certain moves in the AAWD corpus, such as 'credentials' and 'experiential,' had no counts or low counts among the GPT-3.5 Turbo variants, highlighting domain differences compared to the AAWD corpus. The table suggests that human-written arguments are the most argumentatively rich and diverse, with more unique moves across different categories than the generated outputs.

Searching for phrases corresponding to rhetorical moves yielded very low scores across all the auto-generated arguments. Therefore we re-conducted this analysis by prompting GPT-4o to label the counter-arguments generated by GPT-3.5 turbo and human counter-arguments for evidence of all of these rhetorical moves, yet we obtained similar results. The prompts and the LLM-labeled rhetorical moves for our evaluation set are provided in the online repository.

## Automatic Evaluation

Table 7 reports the automatic scores for content and quality for GPT-3.5 finetuned-generated counter-arguments. Table 8 reports the automatic scores for content and quality for Koala-13B generated counter-arguments.

For finetuned Koala-13B, Table 9 reflects the content and style evaluation. In general, we observe that the content and style scores fare poorer than GPT-3.5 turbo. Koala-13B outputs had less content overlap and were less readable than those generated through GPT-3.5 turbo. Koala-13B and Koala-13B-finetuned outputs were also less grammatical, relevant, coherent, and less preferred overall as compared to the counter-arguments generated through GPT-3.5 turbo. The total output and the results for Koala-13B are reported in the Appendix and the supplementary materials[3].

| **Metric** | **Candela** | **GPT-3.5 Turbo No Style** | **GPT-3.5 Turbo Justification** | **GPT-3.5 Turbo Reciprocity** |
|---|---|---|---|---|
| **Automatic Evaluation: Readability (0 to 1 scale)** | | | | |
| Flesch Kincaid Grade | 6.40 (2.18) | **12.81 (2.07)** | 12.75 (2.07) | 11.79 (2.08) |
| Flesch Reading Ease | **83.10 (10.41)** | 40.94 (11.31) | 41.78 (10.62) | 46.23 (11.37) |
| Gunning Fog | 8.85 (2.05) | **15.05 (2.23)** | 15.03 (2.23) | 13.93 (2.17) |
| Smog Index | 8.53 (2.39) | 14.85 (1.89) | **14.87 (1.68)** | 14.09 (1.72) |
| **Rhetorical Intent** | | | | |
| **Move Type** | **Human-written Reddit Counter-argument** | **GPT-3.5 Turbo No Style** | **GPT-3.5 Turbo Justification** | **GPT-3.5 Turbo Reciprocity** |
| **Alignment Moves** | | | | |
| Positive | 12 | 0 | 4 | 2 |
| Negative | 12 | 0 | 4 | 6 |
| **Authority Moves** | | | | |
| Experiential | 10 | 0 | 0 | 6 |
| External | 10 | 0 | 4 | 2 |
| Forum | 10 | 0 | 4 | 4 |
| Social Expectations | 8 | 0 | 0 | 2 |

Table 6: Evaluation of readability and rhetorical intent of different models. The alignment moves identified in Counterfire outputs, based on AAWD features, highlight the discursive richness of human counter-arguments (Bender et al. 2011).

| **Metric** | **Candela** | **FT GPT-3.5 Turbo No Style** | **FT GPT-3.5 Turbo Justification** | **FT GPT-3.5 Turbo Reciprocity** |
|---|---|---|---|---|
| Automatic evaluation: Content (F1 scores) | | | | |
| ROUGE-1 | 0.24 0.24 (0.07) | 0.23 0.24 (0.07) | 0.23 0.24 (0.07) | 0.23 0.23 (0.07) |
| ROUGE-2 | 0.03 0.03 (0.03) | 0.03 0.02 (0.03) | 0.03 0.02 (0.03) | 0.03 0.02 (0.03) |
| ROUGE-L | 0.21 0.21 (0.06) | 0.14 0.14 (0.04) | 0.14 0.14 (0.04) | 0.14 0.14 (0.04) |
| BLEU | 0.00 0.00 (0.01) | 0.01 0.00 (0.02) | 0.00 0.00 (0.02) | 0.00 0.00 (0.02) |
| Automatic evaluation: Style (Debater API) | | | | |
| Evidence support (Pro; Con; Neutral) | 0.99; 0.00;0.00 | 0.96; 0.03; 0.01 | 0.94; 0.02; 0.04 | 0.99; 0.01; 0.00 |
| Argument Quality | 0.54 | 0.76 | 0.46 | 0.63 |
| Automatic evaluation: Readability (0 to 1 scale) | | | | |
| Flesch Kincaid Grade | 6.40 6.00 (2.18) | 12.80 12.25 (5.42) | 12.43 11.55 (5.25) | 12.81 11.05 (6.88) |
| Flesch Reading Ease | 83.10 84.00 (10.41) | 54.18 53.95 (18.32) | 55.24 56.76 (18.54) | 53.99 56.61 (21.67) |
| Gunning Fog | 8.85 8.57 (2.05) | 15.36 14.69 (5.66) | 14.85 14.03 (5.47) | 15.49 13.84 (7.02) |
| Smog Index | 8.53 8.30 (2.39) | 7.55 10.75 (6.82) | 6.80 9.45 (6.55) | 6.77 8.45 (6.58) |

Table 7: Evaluation of the counter-arguments generated by GPT-3.5 turbo fine-tuned reported as the [mean median (standard deviation)].

Table 10 reports the automatic scores for content and quality for Koala-13B-generated counter-arguments.

# Error analysis

## Inspection of Human Evaluation Scores

The examples in Table 11 represent the counterarguments generated by two models that scored among the highest and the lowest on human evaluations of their content quality. Starting with those with the highest scores, the first PaLM-generated counterargument addresses the risks of couchsurfing. It scored a 4.12 in content, which was among the highest scores. This high score correlates with its effectiveness by providing concrete steps to mitigate identified risks, thus presenting a strong counterargument that is both practical and relevant. Similarly, in the second row, the GPT-3.5 fine-tuned model obtained a high score, possibly because it generated many strong arguments on the responsibilities of businesses to provide their workers with a livable wage. In the third row, the PaLM-2 model prompted with justification appears to offer a list of evidence to support its stance, and also scores highly. Note, however, that, unlike the third row, the first two rows do not appear to have adhered to generating reciprocity-style counter-arguments as per their prompt (second column).

The last three rows illustrate counter-arguments with low scores. The fourth row demonstrates that GPT-3.5 fine-tuned models were prone to generate incomplete counterarguments at times, which scored low on content and effectiveness.

[3]https://github.com/Preetika764/Style_control/

| Metric | Candela | Koala No Style | Koala Justification | Koala Reciprocity |
|---|---|---|---|---|
| **Automatic evaluation: Content (F1 scores)** | | | | |
| ROUGE-1 | 0.24 0.24 (0.07) | 0.16 0.17 (0.07) | 0.16 0.17 (0.07) | 0.14 0.15 (0.07) |
| ROUGE-2 | 0.03 0.03 (0.03) | 0.02 0.01 (0.02) | 0.02 0.01 (0.02) | 0.01 0.00 (0.02) |
| ROUGE-L | 0.21 0.21 (0.06) | 0.10 0.10 (0.04) | 0.10 0.10 (0.04) | 0.09 0.10 (0.04) |
| BLEU | 0.00 0.00 (0.01) | 0.00 0.00 (0.01) | 0.00 0.00 (0.01) | 0.00 0.00 (0.00) |
| **Automatic evaluation: Style (Debater API)** | | | | |
| Evidence support (Pro; Con; Neutral) | 0.99; 0.00;0.00 | 0.99; 0.01; 0.00 | 0.99; 0.00; 0.00 | 0.94; 0.04; 0.02 |
| Argument Quality | 0.54 | 0.89 | 0.87 | 0.76 |
| **Automatic evaluation: Readability (0 to 1 scale)** | | | | |
| Flesch Kincaid Grade | 6.40 6.00 (2.18) | 10.68 11.80 (7.26) | 10.69 11.90 (7.11) | 11.97 11.60 (9.69) |
| Flesch Reading Ease | 83.10 84.00 (10.41) | 56.24 48.84 (38.61) | 56.18 48.25 (38.43) | 53.22 48.84 (38.61) |
| Gunning Fog | 8.85 8.57 (2.05) | 13.13 13.62 (4.80) | 13.17 13.78 (4.68) | 14.26 13.44 (7.73) |
| Smog Index | 8.53 8.30 (2.39) | 13.00 14.20 (4.75) | 13.06 14.30 (4.86) | 11.07 13.60 (6.18) |

Table 8: Evaluation of the counter-arguments generated by Koala-13B reported as the [mean median (standard deviation)].

| Metric | Candela | FT Koala No Style | FT Koala Justification | FT Koala Reciprocity |
|---|---|---|---|---|
| **Automatic evaluation: Content (F1 scores)** | | | | |
| ROUGE-1 | 0.24 0.24 (0.07) | 0.25 0.25 (0.09) | 0.25 0.24 (0.09) | 0.25 0.25 (0.09) |
| ROUGE-2 | 0.03 0.03 (0.03) | 0.04 0.03 (0.04) | 0.04 0.03 (0.04) | 0.04 0.03 (0.05) |
| ROUGE-L | **0.21 0.21 (0.06)** | 0.13 0.13 (0.05) | 0.12 0.13 (0.05) | 0.13 0.13 (0.05) |
| BLEU | 0.00 0.00 (0.01) | 0.00 0.00 (0.02) | 0.00 0.00 (0.02) | 0.00 0.00 (0.02) |
| **Automatic evaluation: Style (Debater API)** | | | | |
| Evidence support (Pro; Con; Neutral) | 0.99; 0.00;0.00 | 0.88; 0.05; 0.07 | 0.01; 0.02; 0.87 | 0.69; 0.06; 0.24 |
| Argument Quality | 0.54 | 0.60 | 0.61 | 0.66 |
| **Automatic evaluation: Readability (0 to 1 scale)** | | | | |
| Flesch Kincaid Grade | 6.40 6.00 (2.18) | 6.88 6.50 (3.88) | 6.84 6.40 (4.01) | **6.89 6.50 (3.93)** |
| Flesch Reading Ease | **83.10 84.00 (10.41)** | 74.07 75.61 (17.32) | 73.75 75.40 (19.47) | 74.20 75.76 (18.02) |
| Gunning Fog | **8.85 8.57 (2.05)** | 7.56 6.98 (3.64) | 7.46 6.93 (3.56) | 7.68 7.17 (3.60) |
| Smog Index | 8.53 8.30 (2.39) | 9.03 9.30 (3.22) | **9.10 9.30 (3.21)** | 9.06 9.30 (3.24) |

Table 9: Evaluation of the counter-arguments generated by fine-tuned Koala-13B reported as the [mean median (standard deviation)]. We observe that Koala has about the same content coverage but lower readability than Candela-generated counterarguments. It does not appear to adhere well to the style instructions in the prompts.

| Metric | Candela | PaLM 2 No Style | PaLM 2 Justification | PaLM 2 Reciprocity |
|---|---|---|---|---|
| **Automatic evaluation: Content (F1 scores)** | | | | |
| ROUGE-1 | 0.24 0.24 (0.07) | 0.12 0.12 (0.04) | 0.13 0.13 (0.04) | 0.13 0.13 (0.05) |
| ROUGE-2 | 0.03 0.03 (0.03) | 0.01 0.01 (0.01) | 0.01 0.01 (0.01) | 0.01 0.01 (0.01) |
| ROUGE-L | 0.21 0.21 (0.06) | 0.08 0.09 (0.03) | 0.10 0.10 (0.03) | 0.08 0.08 (0.03) |
| BLEU | 0.00 0.00 (0.01) | 0.00 0.00 (0.00) | 0.00 0.00 (0.00) | 0.00 0.00 (0.00) |
| **Automatic evaluation: Style (Debater API)** | | | | |
| Evidence support (Pro; Con; Neutral) | 0.99; 0.00;0.00 | 0.96; 0.02; 0.02 | 0.97; 0.02; 0.01 | 0.99; 0.00; 0.00 |
| Argument Quality | 0.54 | 0.76 | 0.74 | 0.76 |
| **Automatic evaluation: Readability (0 to 1 scale)** | | | | |
| Flesch Kincaid Grade | 6.40 6.00 (2.18) | 15.07 15.35 (2.62) | 15.90 16.3 (2.78) | 12.53 12.5 (2.21) |
| Flesch Reading Ease | 83.10 84.00 (10.41) | 24.77 23.10 (14.73) | 23.10 23.92 (15.61) | 42.49 46.68 (12.45) |
| Gunning Fog | 8.85 8.57 (2.05) | 16.62 16.62 (2.70) | 17.18 17.98 (3.22) | 13.73 13.77 (2.26) |
| Smog Index | 8.53 8.30 (2.39) | 16.59 16.95 (2.29) | 17.32 17.7 (2.34) | 14.83 14.90 (2.37) |

Table 10: Evaluation of the counter-arguments generated by PaLM 2 reported as the [mean median (standard deviation)].

The last two rows suggest how making repetitive arguments can result in low content quality scores. For instance, the counterargument on language and communication generated by PaLM-2 provides a broad statement on the complexities of language without directly addressing the original claim, which might explain the lower score. Yet, the low content quality score may not necessarily penalize the overall effectiveness of the argument to stay on point.

| Model | Style | Counterargument | Overall | Content |
|---|---|---|---|---|
| PaLM 2 | Reciprocity | **Counterargument:** While it is true that couchsurfing can be a great way to travel and meet new people, there are also some risks involved, such as the potential for sexual harassment or assault. Hosts and surfers should be aware of these risks and take steps to mitigate them, such as communicating openly and setting clear boundaries. Additionally, hosts should be respectful of their surfers' p | 3.5 | 4.12 |
| GPT-3.5 finetuned | Reciprocity | most importantly , it 's really about who should foot the bill to enforce a livable wage .businesses can currently treat their workers poorly because the government has an obligation to its citizens to enforce a livable wage .the issue is n't that we have a too high of a living wage, it 's that business do n't have a thumb to ensure their workers are n't being exploited . | 3.25 | 4 |
| PaLM 2 | Justification | While increasing congressional staffers' salaries may address some concerns about brain drain and lobbyist influence, it is not a comprehensive solution and may have unintended consequences. Here are some counterarguments to consider: 1. **Increased Salaries May Not Solve Brain Drain**: Raising salaries alone may not be sufficient to retain experienced staffers, as many leave for reasons beyond f | 3.5 | 4 |
| GPT-3.5 finetuned | Reciprocity | it depends on who the people and the - what the 's are . | 2.5 | 2.88 |
| PaLM 2 | Justification | **Counterargument:** While it is true that words should serve to promote communication, the argument that language should be changed when it hinders communication is flawed. Language is a complex and nuanced tool that can be used to express a wide range of ideas and emotions. Sometimes, using indirect language or jargon can be more effective than using direct language. For example, when giving neg | 3.38 | 2.62 |
| GPT-3.5 finetuned | Nostyle | i think people do this because they have different ideas of what is fiscally conservative. obviously one of the purest form of fiscal conservatism is just to have as little government as possible , but many people who consider themselves economically conservative do not fall into that category .i 'm making a somewhat arbitrary viewpoint here , but let 's just say that a non-zero amount of government should have a certain level of social safety nets .similarly, i consider myself fiscally moderate ( pure fiscally neutral sounds silly for this aspect ) .i do n't think you should be makin a surplus or a defecit , but a zero overall balance isn't always ideal for business . | 2.62 | 2.38 |

Table 11: Error analysis for human evaluation scores.

## Inspection of ROUGE-L F1 Scores

In Table 12, we analyze counterarguments generated by various models, evaluated on the ROUGE-L F1 metric, which measures the overlap of the generated text with reference texts. Counterarguments from GPT-3.5, PaLM-2, and Koala-13B-finetuned with the highest and lowest scores are included, offering insights into their content quality as perceived through the lens of linguistic similarity.

The GPT-3.5 model's counter-argument on the one-size-fits-all education system received a ROUGE-L F1 score of 0.23, indicating some lexical overlap with reference counterarguments. This argument offers an intricate and well-considered perspective on the topic, with a structured critique and pertinent questioning reflecting the reciprocity style. Similarly, the Koala-13B-finetuned generated counter-argument on ethical egoism holds the highest score in the table at 0.30. The model may have a higher ROUGE-L F1 score due to its use of specialized terminology and philosophical concepts. On the lower end, the PaLM-2 model's justification-style counterargument for the role of the US military includes evidence and a conclusion but perhaps lacks the lexical richness or the direct matching phrases that ROUGE-L F1 scoring favors, hence the lower score.

The last two entries from GPT-3.5-finetuned and GPT-3.5 show particularly low ROUGE-L F1 scores (0.06 and 0.02, respectively), suggesting minimal overlap with reference texts. These arguments, discussing the negative consequences of a society overly focused on comfort and safety, and views on affirmative action, may offer unique or less conventional perspectives, which could explain their low similarity scores. However, this uniqueness doesn't necessarily imply a lack of argumentative quality but may reflect originality in response, which ROUGE-L F1 doesn't capture.

Finally, a counterargument from Koala-13B-finetuned regarding affirmative action also received the lowest score of 0.02, indicating that its content was significantly divergent from the expected responses. This could be due to a unique approach to the topic, a misalignment with the reference texts, potentially highlighting the model's creative angle or a misunderstanding of the topic.

In summary, these examples illustrate that while ROUGE-L F1 scores provide a quantitative measure of linguistic similarity to reference counterarguments, they may not always align with the substantive quality or originality of the content. Models like GPT-3.5 and Koala-13B-finetuned exhibit a range of effectiveness in their counterarguments, with scores reflecting both the ability to engage with conventional discourse and the potential to offer novel perspectives.

| Model | Style | Counterargument | ROUGE-L F1 |
|---|---|---|---|
| GPT-3.5 | Reciprocity | While a one-size-fits-all education system may initially seem like a solution to the issues surrounding teaching and funding, it is necessary to consider the needs and abilities of each individual child. How can a single system cater to the diverse learning styles and abilities of students with different backgrounds and experiences? Is it fair to deprive students of the opportunity to thrive in an environment tailored to their needs? Additionally, implementing such a system may stifle creativity and limit a teacher's ability to adapt their teaching methods to suit the needs of their class. Would it not be better to focus on improving the quality of teaching | 0.23 |
| Koala 13B finetuned | Nostyle | ethical egoism is the normative ethical position that moral agents ought to do what is in their own self-interest .it differs from psychological egoism , which claims that people can only act in their self-interest .ethical egoism also differs from rational egoism , which holds that it is rational to act in one 's self-interest .ethical egoism holds , therefore , that actions whose consequences will benefit the doer can be considered ethical in this sense .<br>/evidence: ["the ethical | 0.30 |
| PaLM 2 | Justification | **Counterargument:** While it is true that the United States military is not constantly engaged in direct combat to defend the freedom of its citizens, it plays a crucial role in maintaining peace and stability in the world, which indirectly contributes to the preservation of freedom.<br>**Evidence 1:** The presence of a strong military deters potential adversaries from initiating conflicts, reducing the likelihood of war and protecting the US from direct threats.<br>**Evidence 2:** The US military engages in peacekeeping and humanitarian missions around the world, promoting stability and preventing the spread of chaos and violence.<br>**Conclusion:** While the military may not be actively fighting on the front lines at all times, its and global presence contribute to the security and freedom of the United States and its allies. | 0.15 |
| GPT-3.5 finetuned | Nostyle | i understand what you 're saying but it 's a moot point for the following reasons : | 0.06 |
| GPT-3.5 | Justification | "While comfort and security may seem like ideal goals, their pursuit can lead to detrimental consequences for individuals and societies. Research shows that the more safety measures a society focuses on, the more it might be creating a weak and ill-equipped generation that struggles to handle inevitable difficulties in life such as death or illness. There is also increasing evidence of rising rates of mental illnesses and suicide among individuals raised in such a system.<br>Moreover, the pursuit of financial and physical safety may actually undermine the society's long-term prosperity. For example, calls for universal basic income, free college education, and raising the minimum wage may" | 0.02 |
| Koala 13B finetuned | Reciprocity | i think the idea of affirmative action is a good one , but i think it should be done in a way that is not discriminatory .i think that the idea of affirmative action is that it is a way to level the playing field for people who have been discriminated against in the past .i think that the idea is that it is a way to give people who have been discriminated against a chance to succeed .i think that the idea is that it is a way to give people who have been discriminated against a chance to succeed .i think that | 0.02 |

Table 12: Error analysis for ROUGE-L F1 scores.

## FActScore

We use FActScore (Min et al. 2023) to evaluate the factuality of our LLM-generated outputs. This measures the factuality of LLM generations by breaking each generation into a series of atomic facts and calculating the percentage of those facts supported by the evidence supplied. The FActScore gives us an assessment of the alignment between generated content and information supplied and suggests that the difference in content between no-style and stylistic variants is about 6-7% on average, where the no-style variant typically has more relevant content. These findings align with our analyses with ROUGE scores. We do observe an anomaly with GPT-4o, where the Reciprocity and Justification style counter-arguments bear up to 22% more content than their no-style counterpart. This suggests that newer LLMs are improving on their ability to integrate fact with style.

| Model | Reciprocity | Justification | No Style |
|---|---|---|---|
| Candela | N/A | N/A | 0.3639 |
| human counterargument | N/A | N/A | 0.6367 |
| GPT-3.5 | 0.5626 | 0.6290 | 0.6243 |
| GPT-3.5 finetuned | 0.4995 | 0.4953 | 0.5659 |
| Koala | 0.5518 | 0.5423 | 0.5390 |
| Koala finetuned | 0.6190 | 0.6418 | 0.6586 |
| GPT-4o | 0.8298 | 0.8792 | 0.6598 |
| Llama-3.1 | 0.7529 | 0.8065 | 0.7687 |
| Haiku | 0.8737 | 0.9114 | 0.8872 |

Table 13: FActScore values of the LLMs for 3 styles

## Limitations

### Data and Context Constraints

We focused on evaluating the style and quality of the counter-arguments generated by LLMs, particularly in terms of their alignment with justification and reciprocity, two rhetorical styles grounded in political communication theory. Our findings should be interpreted in cognizance of the limitations of our analysis. Firstly, the evidence sources for argument generation include Wikipedia, Washington Post, New York Times, Reuters, and The Wall Street Journal. While we acknowledge that these sources may have varying political leanings and editorial stances. The source distribution served our research purpose by providing a consistent foundation across all experimental conditions. This allowed us to isolate the impact of stylistic variations while holding the evidence constant. Related to this, while we assumed that the fact retrieval system adapted from Hua, Hu, and Wang (2019) was functioning as intended, our analysis was limited to English political posts due to the constraints of the Candela dataset. We would welcome future research expanding our work to other languages, as well as examining how different evidence sources might interact with stylistic choices in counter-argument generation. Secondly, the dataset was co-created by political users on Reddit, familiar with a set of social norms typical of the r/CMV subreddit. Our dataset, derived from ChangeMyView, reflects a context where users actively engage with those holding opposing views. Therefore, the data characteristics may be complex to replicate even when a general population of Reddit users is familiarized with the rules of r/CMV and invited to participate in a political debate using the same experimental conditions. Thirdly, while the effectiveness of different arguments may differ in a real-life or face-to-face political discussion, our evaluation of human preference offers at least a basic understanding of the kind of counter-arguments that humans prefer to read in a discussion forum, such as Reddit.

## Model and Evaluation Challenges

In terms of operational limitations, firstly, given the resource-intensive nature of fine-tuning, our experiments with GPT-3.5 Turbo and GPT-4o involved API calls that resulted in high latency. Secondly, because models available through APIs are sometimes phased out, our results may cease to be replicable a few down the line. For instance, because LLMs are increasingly released with guardrails against controversial text generation, we may not be able to produce counter-arguments for our prompts in the future (and were not always to generate them during our experiments). In particular, LLaMA-3.1 did not always produce outputs, citing ethical concerns. We also observed that Claude's tendency to prioritize neutrality results in less forceful counterarguments, and it occasionally refused to provide counterarguments due to ethical considerations. We have documented all the prompts used in the different LLMs in the online repository, together with the corpus.

Thirdly, although our choice of evaluation metrics aligns with established practices, particularly regarding style adherence and factuality, these metrics have inherent limitations. Metrics such as ROUGE and BERTScore have limitations. As discussed in the Error Analysis section, they may fail to capture creative or unique perspectives and are susceptible to misalignment with reference texts. Finally, this analysis compared LLM performance using consistent prompts across all models, leaving prompt sensitivity unexplored—an area that warrants future investigation. Therefore, we have highlighted the specific challenges LLMs face in generating coherent and contextually appropriate counter-arguments. For instance, not all LLMs adhered to our prompts. PaLM-2 produced inconsistencies, generating supportive arguments in approximately 10% of cases rather than counter-arguments, which led us to exclude it from fine-tuning. As another example, despite incorporating external evidence, GPT models continue to exhibit biases and hallucinations. Our future experiments will explore fine-tuning LLMs for quality-specific tasks and broader argumentation datasets.